\documentclass[10pt,onecolumn,letterpaper]{article}
\usepackage{graphicx}
\usepackage{xcolor}
\usepackage{multirow}
\usepackage{graphicx}
\usepackage{longtable}
\usepackage{csvsimple}
\usepackage{bbold}
\usepackage{amsmath}
\usepackage{amssymb}
\usepackage{hyperref}
\usepackage{url}

\newcommand\Tstrut{\rule{0pt}{2.6ex}}
\newcommand\Bstrut{\rule[-1ex]{0pt}{0pt}}

\newcommand{\E}{\mathbb{E}}
\newcommand{\R}{\mathbb{R}}

\begin{document}

\title{Segmenting the motion components of a video: A long-term unsupervised model}

\author{Etienne Meunier and Patrick Bouthemy \\~\\
Inria, Centre Rennes - Bretagne Atlantique, France}

\maketitle

\begin{abstract}
Human beings have the ability to continuously analyze a video and immediately extract the motion components. We want to adopt this paradigm to provide a coherent and stable motion segmentation over the video sequence. In this perspective, we propose a novel long-term spatio-temporal model operating in a totally unsupervised way. It takes as input the volume of consecutive optical flow (OF) fields, and delivers a volume of segments of coherent motion over the video. More specifically, we have designed a transformer-based network, where we leverage a mathematically well-founded framework, the Evidence Lower Bound (ELBO), to derive the loss function. The loss function combines a flow reconstruction term involving spatio-temporal parametric motion models combining, in a novel way, polynomial (quadratic) motion models for the spatial dimensions and B-splines for the time dimension of the video sequence, and a regularization term enforcing temporal consistency on the segments. We report experiments on four VOS benchmarks, demonstrating competitive quantitative results, while performing motion segmentation on a whole sequence in one go. We also highlight through visual results the key contributions on temporal consistency brought by our method.
\end{abstract}

\newcommand{\fix}{\marginpar{FIX}}
\newcommand{\new}{\marginpar{NEW}}

\maketitle

\section{Introduction}
\label{sec:intro}

When dealing with videos, motion segmentation is one of the key issues. Human beings have the ability to continuously analyze a video and immediately extract the diverse motion components.
We want to design a similar paradigm to provide a coherent and stable motion segmentation over the video sequence. 
Since optical flow (OF) yields all the information on the movement between two images of the video sequence, it is natural to base motion segmentation on optical flow.
Motion segmentation is a computer vision task in its own, but it is also useful for downstream tasks.
Indeed, it is essential to have an accurate and reliable motion segmentation for independent moving object detection, multiple object tracking, or video understanding, to name a few. Motion segmentation based on optic flow has recently received renewed attention in the context of deep learning \cite{meunier2022,xie2022,yang_motion-grouping_2021,yang_unsupervised_2019}, but the temporal extension remains to be further investigated.

In this paper, we propose an original long-term unsupervised model for segmenting all the motion components of a video, while taking optical flow as input. It must therefore be considered first and foremost as an unsupervised method performing optical flow segmentation (OFS) over an entire sequence. We assume that optical flow is readily available, computed by one of the recent efficient optical flow methods \cite{tu2019}. The optical flow field at time $t$ enables to get the motion segmentation map at frame $t$ of the video.
However, taking a large time window, and even the whole video sequence, is beneficial, since motion entities are generally consistent throughout a video sequence (or at least within every video shot for long videos).
Indeed, temporal coherence is inherent to motion.

Motion components (or segments) observed in an image sequence may come from a variety of sources. The first that springs to mind is the case of one or several independent moving objects in the viewed scene. In case of articulated motion, a motion segment will ensue from each part of the articulated body. If the camera is itself moving, the static scene exhibits an apparent motion in the video, which may induce several motion segments according to the depth and/or orientation of the static objects in the scene. In particular, a static object in the foreground of the scene will have an apparent motion different than the scene background one, which is usually referred to as motion parallax \cite{irani1998}.

Even if we take an object-oriented view of video content, motion appears as an intrinsic, globalizing property of an object. It is more immediate to segment an object based on its motion than on its appearance. Indeed, motion has been used in the training stage to facilitate image segmentation at inference time \cite{choudhury2022}, or as a criterion to validate image segments \cite{mahendran2018}. Motion segmentation is also frequently involved in video object segmentation (VOS).

Admittedly, better performance is obtained in VOS, if appearance and motion are combined, compared to the single use of motion. However, there is a bias in the VOS datasets: the primary object to be detected is not only salient by its motion compared to the background, but also by its appearance and even its location, the object being generally placed in the image center. Yet, combining the two modalities may confine their use to videos where the appearance of the object of interest can be easily represented by deep features, like DINO features \cite{caron2021} that are trained on a huge set of natural images from the Internet. Indeed, such algorithms could not be easily applied to videos depicting other types of objects and contents, such as satellite images, biological images, underwater or industrial videos. Models trained on optical flow have a much better generalization and domain transfer ability. %
Moreover, relying on motion allows to deal with contents where appearance is not discriminating, for instance for crowd scenes \cite{meunier2021}, or for moving camouflaged animals in natural scenes \cite{lamdouar_camouflage_2020}.

Our long-term unsupervised spatio-temporal model is a significant extension of the model we introduced in \cite{meunier2023-cvpr}. 
For the sake of clarity, from now on, we will name the method presented in this paper LT-MS (for long-term motion segmentation) and the previous one described in \cite{meunier2023-cvpr} ST-MS (for short-term motion segmentation).

First, we have designed a fully end-to-end method, whereas ST-MS required an \textit{ad-hoc} post-processing step to achieve the temporal linkage beyond three successive frames. This was made possible by defining a new space-time parametric motion model to handle more elaborated temporal evolutions. The spatial part of the motion model remains a polynomial model (e.g., a quadratic one), while the temporal part results from a B-spline approximation. Besides, we have added a transformer (only the decoder part) on the latent space of the 3D U-net, allowing for long-range interactions. While ST-MS takes only three consecutive flows as input, LT-MS can accommodate at inference time any length of series of flow fields, ten, twenty, one hundred, or even the whole sequence.
Second, the LT-MS loss function is similar to the ST-MS one, but it is now inferred in a mathematical well-founded manner, leveraging the Evidence Lower Bound (ELBO) framework. We also propose a mathematical development to motivate the temporal consistency term of the loss function. Third, we have added a fourth dataset in the experimental evaluation, the DAVIS2017-motion dataset introduced in \cite{xie2022}.

The rest of the paper is organized as follows. Section \ref{sec:related-work} describes related work regarding motion segmentation. Section \ref{method} presents our unsupervised network for long-term multiple motion segmentation. In Section \ref{implementation}, we provide implementation details. Section \ref{results} reports results on four VOS benchmarks with a comparison to state-of-the-art unsupervised motion segmentation methods. Finally, Section \ref{conclusion} contains concluding remarks.

\section{Related work}
\label{sec:related-work}

Motion segmentation aims to break down each frame of a video sequence into components (or segments) of coherent motion. Usually, each motion segment is identified by a motion model, which can be hand-crafted such as affine or quadratic polynomial models, or sometimes learned \cite{tokmakov_learning_2017}.
Motion segmentation has been investigated for decades \cite{cremers2005,mattheus2020,mitiche1996,sun2012,wang1994,xiao-shah2005,zappella2008}. However, we will focus on recent methods in this section. We refer the reader to \cite{meunier2022,mitiche1996,zappella2008} for a review of older works.
Since accurate and efficient methods for estimating optical flow are now available, motion segmentation methods can leverage optical flow as a reliable input. The advent of deep learning, and more recently the use of transformers with attention mechanisms \cite{han2023}, has also encompassed motion segmentation.

In \cite{yang_motion-grouping_2021}, a transformer module, more specifically, the slot attention mechanism introduced in \cite{locatello2020}, is leveraged to perform motion segmentation from optical flow. As a matter of fact, it addresses a binary segmentation problem, foreground moving object \textit{vs} background. The loss function is composed of two terms, a flow reconstruction term and an entropy term to make masks as binary as possible. In \cite{bideau_moa-net_2018}, it is assumed that camera rotation can be removed in the optical flow field, which leads to a simplified scheme to detect independently moving objects in a self-supervised way. Another approach is adopted in \cite{xu2021} that is able to cope with multiple motion segmentation. Nonlinear subspace filters are learned from stacked deep multi-layer perceptrons. Then, motion segmentation is inferred at inference by applying $K$-means to the output embeddings. In \cite{meunier2022}, we leveraged the Expectation-Maximization (EM) framework to define the loss function and the training procedure for an unsupervised frame-by-frame motion segmentation, where 12-parameter quadratic motion models are involved. 
In \cite{yang_unsupervised_2019}, the authors developed an adversarial method whose aim is to generate a mask hiding the input optical flow, where an inpainter network attempts to recover the flow within the mask. The rationale is that no correct reconstruction of the inside flow can be achieved from the outside flow, if the hidden area exhibits an independent motion and then constitutes a motion segment.

The temporal dimension of motion segmentation has already been considered in different ways. Regarding classical approaches, in \cite{odobez1995} the motion partition at time $t$ was predicted from the one obtained at time $t-1$ using the affine motion models estimated between images $t-1$ and $t$ for each segment, within a robust MRF-based method. The authors in \cite{sun2012} showed that it was beneficial to introduce temporal layer constancy over several frames to perform motion segmentation in a MRF-based and graph-cut optimization framework. In \cite{ochs2014}, large time windows were taken into account, but point trajectories were used as input within a spectral clustering method.

\noindent
Regarding deep learning approaches, motion segmentation at time $t$
is improved at training time in \cite{yang_motion-grouping_2021}, by taking, as input, flows between $t$ and several time instants before and after $t$.
The authors of \cite{ding2022} consider several consecutive RGB frames as input of their self-supervised method. Optical flow is only computed at training time, and the loss function also comprises a temporal consistency term. However, the latter is not applied to two consecutive segmentation masks, but for pairings between the same frame $t$ and another (more or less distant) one. In \cite{duke2021}, spatio-temporal transformers were designed for video object segmentation involving temporal feature propagation. In \cite{ye2022}, the authors introduce deformable sprites with a video auto-encoder model, but optimized at test time, to segment scene elements through the video. It leverages texture and non-rigid deformations. The work \cite{ponimatkin2023} involves pretrained appearance features (using DINO) and optical flow. The objective function is derived from a form of spectral clustering and applied to the entire video. However, optimization is required at test time. In \cite{meunier2023-cvpr}, we explicitly handled the temporal dimension of the problem by considering triplets of consecutive flows as input, a space-time polynomial motion model, and a temporal consistency term in the loss function. 

Since motion segmentation can be used in the context of VOS and that we will comparatively evaluate our method on VOS benchmarks, we briefly review VOS methods. The latter often combine appearance and motion features \cite{cheng2017,griffin_tukey-inspired_2019}. It was addressed with supervised or semi-supervised methods as in \cite{dave2019,duke2021,jain_fusionseg_2017,dystab2021}, but also with unsupervised methods \cite{meunier2022,yang_motion-grouping_2021,yang_unsupervised_2019}. VOS is concerned with the segmentation of primary objects, i.e., object moving in the foreground of a scene and possibly tracked by the camera \cite{wang_survey-VOS}. Thus, VOS generally involves binary ground truth segregating primary moving object against background. This is for instance the case for the DAVIS2016 benchmark \cite{perazzi_benchmark_2016}.
Recent works have revisited the way of coupling appearance and motion for VOS. The AMD method \cite{liu2021} includes two pathways, the appearance one and the motion one.
If it does not use optical flow as input and brings out the objectness concept, it nevertheless relies on a coarse motion representation.
The RCF method \cite{lian2023} involves learnable motion models, and is structured in two stages, a motion-supervised object discovery stage, and a refinement stage with residual motion prediction and high-level appearance supervision. However, the method cannot distinguish objects undergoing different motions. In \cite{choudhury2022}, the prediction of motion patterns is used at training time to learn objectness from videos.
The DivA method \cite{lao2023} promotes divided attention. It is based on the same principle as in \cite{yang_unsupervised_2019} that motion segments are mutually uninformative, but it is not limited to binary segmentation.
It can segment a variable number of moving objects from optical flow, 
by leveraging a slot attention mechanism guided by the image content through a cross-modal conditional slot decoder.

Our fully unsupervised long-term approach differs from these previous works in several respects as outlined in the introduction.

\section{Long-term Motion Segmentation Method}
\label{method}

\begin{figure*}[tbh!]
\centering
\includegraphics[width=0.90\linewidth]{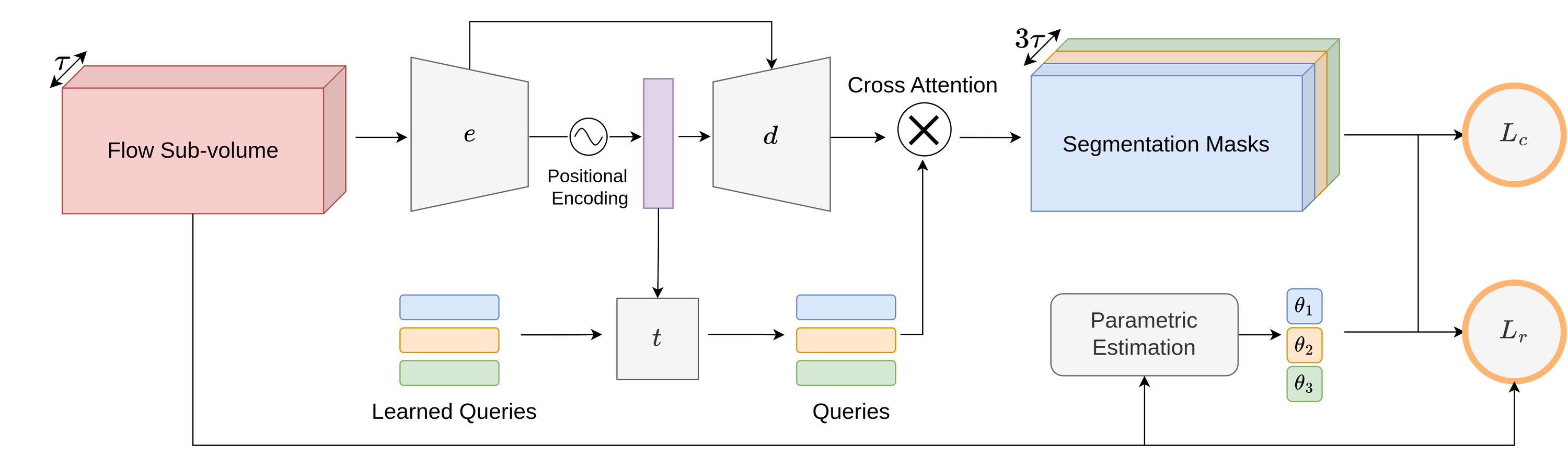}
\caption{Overall architecture of our multiple motion segmentation method ensuring temporal consistency with the loss term $\mathcal{L}_c$ and the B-spline space-time motion models $\theta_k$ (for $k=1,..,K$). It takes as input a volume of $T$ flow fields. It comprises a 3D U-net ($e$ and $d$ boxes) and a transformer decoder ($t$ box). It also involves positional encoding. A cross-attention product yields the $K$ segmentation masks corresponding to the input volume
For the sake of clarity, the block diagram is represented for three motion segments ($K=3$). $\mathcal{L}_r$ is the flow-reconstruction loss term.}
\label{overall-architecture}
\end{figure*}

We have designed a transformer-based network for multiple motion segmentation from optical flow. It is inspired from the MaskFormer architecture \cite{cheng2021}, but it only comprises one head corresponding to the mask prediction, as described in Fig.\ref{overall-architecture}. The network takes as input a volume, of flexible temporal length, comprising several consecutive optical flow fields. Temporal consistency is expressed in two main ways at the training stage. Firstly, we associate a space-time motion model with each segment to characterize its motion along with the evolution of the latter over time. Secondly, the loss function comprises an additional term  enforcing consistent labeling of the motion segments over the volume. Adding the transformer to the latent space of the 3D U-net allows for interaction between distant features in the input, and thus for long-term segmentation. We only incorporate a transformer decoder on the downsampled latent space, which induces a limited additional computational load.

\subsection{Spatio-temporal parametric motion model}
\label{motion-model}
The set of $T$ consecutive flows will be designated as a space-time volume (or volume, to make it short). The volume could even be the whole video sequence. Our space-time motion model is estimated through B-spline functions \cite{unser1999} for its temporal dimension.
More precisely, we consider a B-spline function of order $n$ in the variable $t$ over the space-time volume.
For the spatial extension, we assign a parametric motion model $\tilde{f}_{\theta_k}$ to each motion segment $k, k=1,..,K$, where $\theta_k$ specifies the spatial motion model $\tilde{f}$ for segment $k$.
In practice, we take $n=3$, which proved sufficient to handle motion evolution. The number $L$ of control points is given by $L=2+\lfloor \frac{T-2}{\nu} \rfloor$, where
$\nu$ manages the temporal frequency of control points. We put a control point at both ends of the volume ($t=1$, $t=T$), and the other control points are equidistantly placed between the two. The control points are often not located at time points of the video.

\begin{figure*}[tbh!]
\centering
\includegraphics[width=0.84\linewidth]{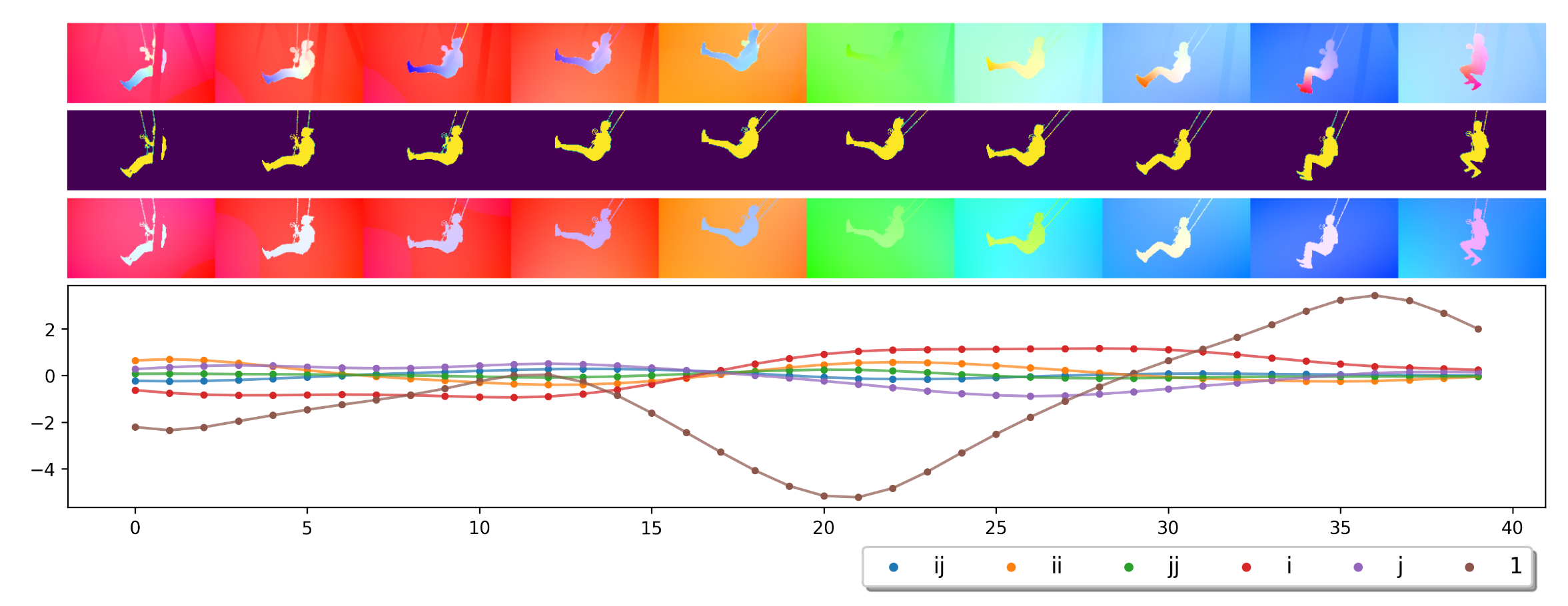}
\caption{Illustration of the spatiotemporal spline-based motion model. Top group: respectively, input flows displayed with the HSV code for the \textit{swing} video of DAVIS2016, binary segmentation ground truth, flows generated by the estimated spline-based motion models for the two segments. Bottom row: plot of the temporal evolution of the six estimated motion model parameters corresponding to the horizontal flow component for the foreground moving object. The $x$-axis is the video frame index.
}
\label{spline-motion}
\end{figure*}

The space-time spline-based motion model is illustrated in Fig.\ref{spline-motion}. Just for this illustration and for the sake of clarity, the motion models are computed within the two segments, foreground moving object and background, provided by the ground truth. The estimated motion model for the foreground moving object is able to capture the periodic nature of the swing motion as demonstrated by the plots of the computed motion model parameters. Also, the motion model computed in the background segment perfectly fits the camera motion. Articulated motion (woman's legs) would require multiple-motion segmentation.

Any parametric motion model could be considered. We take a polynomial expansion in the coordinates $x$ and $y$ of any point in the image, with one polynomial for each component of the 2D flow. The use of polynomial is a usual choice for 2D motion modelling \cite{mitiche1996}, since it is easy to estimate and enables a straightforward physical interpretation. More specifically, we select the 12-parameter quadratic motion model to be able to account for continuously varying depth surface of the objects in the scene, especially for the whole background, and for complex object or camera motions. In contrast, the affine and the 8-parameter quadratic motion models assume a planar object surface. Indeed, the latter exactly corresponds to the projection in the image of the 3D rigid motion of a planar surface. It is equivalent for velocity fields to the homography transform for points. However, in presence of severe depth effects (strong depth discontinuities) and camera motion, the static scene cannot be represented by a single motion model due to motion parallax produced by static objects located in the foreground.

Regarding first the spatial dimension of the motion model, the 2D flow vector yielded by the full quadratic motion model at point $(x,y)$ writes:
\begin{align}
\label{12quadratic}
\tilde{f}_{\theta}(x,y)=&(\theta_{1}+\theta_{2}x+\theta_{3}y+\theta_{7}x^2+\theta_{8}xy+\theta_{9}y^2, \nonumber\\
 &\theta_{4}+\theta_{5}x+\theta_{6}y+\theta_{10}x^2+\theta_{11}xy+\theta_{12}y^2)^T.
\end{align}
B-Splines are modelling the evolution of the parameters of the motion model over time. The resulting motion model at a given time is then a weighted average of the close control ones, each control one being a quadratic model as in eq.(\ref{12quadratic}). By denoting now $S_n(\theta_k)$ the subscript of $\tilde{f}$, we emphasize that the motion model parameters for each segment $k$ are estimated through the B-spline functions:
\vspace{-0.18cm}
\begin{equation}
\label{b-spline}
\tilde{f}_{S_n(\theta_k)}(x,y,t) = \sum_{l=1}^L \tilde{f}_{\theta_{k,l}}(x,y) B_{n,l}(t),
\end{equation}
\vspace{-0.05cm}
where $B_{n,l}$ is the $l^{th}$ B-spline, $\theta_{k,l}$ corresponds to the $l^{th}$ control point of the spline, and $\tilde{f}_{\theta_{k,l}}$ is defined by eq.(\ref{12quadratic}).

\subsection{Loss function}

We consider a volume of optical flow fields $f \in \R^{2 \times T \times W \times H}$ on the image grid $\Omega \in \R^{W \times H}$ of width $W$ and height $H$, and over $T$ temporal steps (or, in other words, a sequence of length $T$).  We denote $f(i,t) \in \R^2$  the flow associated to site (or point) $i\in \Omega$ ($i=(x,y)$) at time $t$. We assume that we can decompose the flow as a set of $K$ segments, each one exhibiting a coherent motion. Flow vectors within a given segment $k$ are represented by a smooth parametric motion model parametrized by $\vartheta_k$. Variable $z_{i,t}$ conveys the motion segmentation, $z_{i,t}^k = 1$ if site $(i,t)$ belongs to segment $k$. $z=\{z_{i,t}^k,i\in\Omega,t=1,\cdots T, k=1\cdots K\}$ and $\vartheta=\{\vartheta_k,k = 1\cdots K\}$ are the latent variables, and $f$ is the observed data. 
Following \cite{kingma2019}, we introduce an approximate distribution over segmentation and motion model parameters:
\begin{equation}
    q(z, \vartheta | f) = q(z|f)q(\vartheta) =  (\prod_{t=1}^T \prod_{i \in \Omega} \prod_{k=1}^Kq(z_{i,t}^k | f))(\prod_{k}^K q(\vartheta_{k})),
\end{equation}
where $q(\vartheta_k) \triangleq \delta(\theta_k)$ with $\delta$ the Dirac distribution and $q(z_{i,t}^k |f) = g_{\phi}(f)_{i,t}^k$. $g_\phi$ is our network model taking as input the optical flow volume and returning a probabilistic segmentation volume. 
We can also write the data-likelihood over a dataset $\mathcal{D}$ of optical flow volumes $f$ as: 
\begin{align}
    \log(\mathcal D) = & \sum_{f\in D} \log p(f) \nonumber \\ =& \sum_{f\in D} \E_{q(z, \vartheta|f)}[\log \frac{p(f, z, \vartheta)}{q(z, \vartheta |f)}] + KL(q(z,\vartheta|f)||p(z, \vartheta | f))  \nonumber \\
    \geq& \sum_{f\in D} \E_{q(z, \vartheta|f)}[\log p(f | z, \vartheta)]  -  KL(q(z, \vartheta |f) || p(z, \vartheta)),
    \label{ELBO}
\end{align}
where we get the Evidence Lower Bound (ELBO) \cite{kingma2019}.

Considering that the flow field can be represented by segment-wise parametric motion models, and assuming conditional independence, we have: 
\begin{align}
    p(f| z, \vartheta) &= \prod_{t=1}^T \prod_{k=1}^K p(f(t) | \vartheta_k, z) \nonumber \\ 
    &\propto \prod_{t=1}^T \prod_{k=1}^K \prod_{i\in \Omega} \exp(-||f(i,t) - \tilde{f}_{S_n(\vartheta_k)}(i,t)||_1 /  \xi_{f, t})^{z_{i,t}^k},
\end{align}
where $\tilde{f}_{S_n(\vartheta_k)}(i,t)$ is the flow vector given by the parametric motion model of parameters $\vartheta_k$ at point $(i,t)$ as defined in Section \ref{motion-model} through the spline approximation and $\xi_{f,t} = \sum_{i\in\Omega} ||f(i,t)||_1$ is a normalizing factor. 
From this likelihood, we can write the left term of the ELBO as: 
\begin{align}
\label{reconstruction}
    \mathcal{L}_r(f) &= - \E_{q(z, \vartheta|f)}[\log p(f|z, \vartheta)] = - \sum_{t,k} \E_q[\log p(f(t) | \vartheta_{k}, z)]
    \nonumber \\& = \sum_{t=1}^T \sum_{i\in \Omega} \sum_{k=1}^K  g_{\phi}(f)_{i,t}^k ||f(i,t) - \tilde{f}_{S_n(\theta_k)}(i,t)||_1 / \xi_{f, t}
\end{align}
The term $KL(q(z,\vartheta |f) || p(z,\vartheta)))$ in eq.(\ref{ELBO}) allows us to define a prior over the segmentation. We want to enforce the property that the segmentation labels are temporally consistent, i.e., similar between neighbouring frames. More formally, we express a prior on $z$ as follows, assuming point independence: 
\begin{align}
        &\forall i \in \Omega, \forall t \in \{1 \cdots T \}, z_{i,t} \in \{1\cdots K\} \nonumber \\
    &p(z) = \prod_{i\in\Omega} p(z_{i,1})\prod_{t=2}^T p(z_{i,t} | z_{i,t-1}), \nonumber \;\; \mbox{with} \\
  &p(z_{i,t} = k | z_{i,t-1} = l) = \frac{1}{M} \exp(\alpha \mathbb{1}[k= l] + \beta \mathbb{1}[k \neq l])\\
  & \mbox{and} \; \; p(z_{i,1} = k) = \frac{1}{K},
\end{align}
where $M$ is a normalizing factor. We introduce the more compact notation $g_{i,t}^k = g_{\phi}(f)_{i,t}^k$.
Using an uninformative prior on $\theta$, the KL term is given by: 
\begin{align}
    KL[q(z,\theta|f) || p(z, \theta)] &= KL[q(z|f) || p(z)] \nonumber \\
    &= \E_{q}[\log q(z|f)] - \E_{q}[\log p(z)].
\label{KL}
\end{align}
We then infer the two terms of eq.(\ref{KL})
\begin{align}
    \E_{q}&[\log q(z|f)] = \nonumber \\
    &\sum_{i\in\Omega} \sum_{t=2}^T \sum_{k=1}^K \sum_{l=1}^L g_{i,t}^kg_{i,t-1}^l (\alpha \mathbb{1}[k=l] + \beta \mathbb{1}[k \neq l]) + c,
\end{align}
where $c=-|\Omega| (\log K+ T\log M)$ is a constant. 
We set $\beta=0$, and knowing that $\alpha >0$, we have: 
\begin{equation}
    \E_{q}[\log p(z)] = \alpha \sum_{i\in\Omega}\sum_{t=2}^T \sum_{k=1}^K g_{i,t}^kg_{i,t-1}^k + c.
\end{equation}
Since 
$2*g_{i,t}^kg_{i,t-1}^k=-(g_{i,t}^k- g_{i,t-1}^k)^2 + (g_{i,t}^k)^2 + (g_{i,t-1}^k)^2$,
after simple calculations, we get: 
\begin{align}
    KL&[q(z | f) || p(z)] = \frac\alpha 2  \sum_{i\in\Omega}\sum_{t=2}^T\sum_{k=1}^K (g_{i,t}^k - g_{i,t-1}^k) ^2 
    \nonumber \\
    &+ \sum_{i\in\Omega}\sum_{t=1}^T\sum_{k=1}^K g_{i,t}^k*(\log(g_{i,t}^k)  - \frac \alpha2*g_{i,t}^k) \nonumber \\
    &-\frac\alpha 2 
    \sum_{i\in\Omega}\sum_{t=2}^{T-1}\sum_{k=1}^K (g_{i,t}^k)^2)  + c.
\end{align}
Since $\forall x\in[0, 1],
-\frac\alpha 2 \leq x(\log x - \frac\alpha 2 x) \leq 0$, we can infer: 
\begin{align*}
    KL[q(z | f) || p(z)] &\leq \frac\alpha 2  \sum_{i,t,k} (g_{i,t}^k - g_{i,t-1}^k) ^2  + c 
\end{align*}
As $\forall x,y \in [0,1]^2, |x-y| \geq (x-y)^2$, we end up with the term of temporal consistency in the loss function  we introduced in ST-MS \cite{meunier2023-cvpr}.
Thus, we have provided a mathematical development to motivate it.
More precisely, it writes:
\begin{align}
\label{consistency}
    \mathcal{L}_c(f) = \frac{1}{2K|\Omega|} \sum_{i\in \Omega} \sum_{t=2}^{T} 
     \sum_{k=1}^K \; |g_{\phi}(f)_{i,t}^k -g_{\phi}(f)_{i,t-1}^k|.
\end{align}
The Eulerian temporal constraint expressed by eq.(\ref{consistency}) does not mean that the site
should be static between two successive frames, but it holds as long as the site lies within the successive projections of a given moving part so that it remains in the same mask. However, it does not make sense on occluded or disoccluded areas. This justifies the use of the L1 norm to deal with the latter configuration.
We further prevent from enforcing the temporal consistency over occlusion areas by ignoring sites $i$ in the summation over $\mathcal{I}$ in eq.(\ref{consistency}) that exhibit a large temporal flow difference. More precisely, for each input flow sequence, we set a threshold $\lambda$ so that a quantile $\eta$ of sites $i$ is discarded as follows:
\begin{equation}
p(\|f(i,t+1)-f(i,t)\|_1 \ge \lambda) \le \eta.
\label{occlusion}
\end{equation}
In practice, we take $\eta=1\%$ with an empirical distribution for $p$ in eq.(\ref{occlusion}). In doing so, we make an implicit assumption on the overall surface of the occlusion areas, but it seems reasonable for the datasets we deal with.

The loss function is finally defined by: 
\begin{equation}
\label{loss-function}
     \sum_{f \in \mathcal D} \mathcal{L}(f,\phi,\theta) = \sum_{f \in \mathcal D} \mathcal{L}_r(f, \phi, \theta) + \gamma \mathcal{L}_c(f, \phi).
\end{equation}
We set $\gamma=1$. The training procedure alternatively updates $\theta$ and $\phi$ with $\epsilon$ as learning rate: 
\begin{align}
\label{it-optimization}
\text{for } f\in D: \; \;  & \theta^* = \arg \min_{\theta} \mathcal{L}(f,\phi,\theta)(f) \; ; \;
 \\ 
& \phi_{t+1} =  \phi_t - \epsilon \nabla_{\phi} \mathcal{L}(f,\phi,\theta^*)
\end{align}

\subsection{Network architecture}
The overall architecture of our unsupervised multiple motion segmentation framework is illustrated in Fig.\ref{overall-architecture}. It includes two main modules. The first one, taking the flow volume as input, is a 3D U-net \cite{ronneberger_Unet_2015}. The latent code augmented with embedding positions is directed to the transformer decoder. Then, by cross-attention, the output formed by the volume of segmentation masks is produced. The training of the overall architecture is based on the minimization of the loss function defined in eq.(\ref{loss-function}), while the motion model parameters of the segments are given by the B-spline estimation. Temporal consistency is provided by the loss function and the space-time motion models.

\section{Implementation}
\label{implementation}

\subsection{Implementation details}
Following \cite{choudhury2022,lian2023,meunier2022,meunier2023-cvpr,yang_motion-grouping_2021}, we adopt the RAFT method \cite{teed_raft_2020} to compute the optical flow fields. More specifically, we use the RAFT version trained on the MPI Sintel dataset \cite{butler2012}. We downsample the computed flow fields to feed the network with 128$\times$224 vector fields. The output segmentation maps are upsampled to the initial frame size for evaluation. Thus, we can perform efficient training and inference stages. 
We typically take flow volumes of temporal length $T=9$ at training time. However, at test time, we can process flow volumes of larger temporal length, and even the flow volume of the whole sequence. To compute the spatio-temporal parametric motion models, $x$ and $y$ coordinates are normalized within $[-1,1]$, and a similar normalization is applied to the $t$ coordinate.
The full 12-parameter quadratic motion model is used in all the experiments. We take $\nu=3$ for the frequency factor in the B-spline approximation.
We select for each point $(x,y)$ the segment $\hat{k}$
with the highest prediction.
In eq.(\ref{consistency}), we set $\lambda$ as the $99^{th}$ quantile of the flow differences over the flow volume. 

We use Adam optimizer with the following strategy on the learning rate $\epsilon$ to train the network (3D U-net and transformer), inspired from the warmup-decay strategy \cite{xie2022}. We linearly increase it from $0$ to $1e-4$ for 20 epochs, then, we divide it by two every 40 epochs. 
The estimation of the motion model parameters through the B-spline approximation at training time is achieved with the Pytorch implementation of L-BGFS \cite{lbgfs89}. Indeed, the parametric models induced over time by the splines are a linear combination of quadratic parametric models (see eq.(\ref{b-spline})). Thus, we can optimize eq.(\ref{reconstruction}) to obtain the argmin in eq.(\ref{it-optimization}), which is a convex problem.

Our LT-MS method is very efficient at test time (code made available upon publication). The computational time amounts on average to 210 \textit{fps} on a GPU P100, that is, twice as fast as the ST-MS method \cite{meunier2023-cvpr}. It is due to the long flow sequence given as input to the network, which allows for parallelisation of some heavy computations. In addition, our LT-MS architecture remains lightweight, since it combines only three U-net layers and a transformer decoder on the downsampled feature space. Let us also stress that our LT-MS method does not involve any post-processing at all.

\subsection{Data augmentation and network training}
We introduced two types of data augmentation dedicated to optical flow. First, we add a 2D global flow to the input flow similarily as we did in the EM \cite{meunier2022} and ST-MS \cite{meunier2023-cvpr} methods.
However, for LT-MS, the 2D global flow is given by a full spline-based spatio-temporal motion model whose parameters are chosen at random.
The same 2D global flow is added to the flow fields of a given input volume. It allows us to mimic diverse camera motions, enforcing that the motion segments are independent of the 2D added global flow, as justified in \cite{meunier2022}. In addition, we corrupt a few flows out of the nine input ones. Thus, we simulate poorly estimated flow fields at some time instants, and the temporal consistency should overcome it.

Our motion segmentation method is fully unsupervised. We never use any manual segmentation annotation.
We train our model using the FlyingThings3D (FT3D) dataset \cite{ft3d2016}.
Our (demanding) objective is to demonstrate the high generalization power of our model.
We train our model once for all on the optical flow fields computed on FT3D, and then, apply it on new unseen datasets.
For hyperparameter setting, we select the stopping epoch from the loss function evaluated on the DAVIS2016 training set.
We trained on a single Quadro RTX 6000 for 77h with a batch size of 5 on 32k flow fields computed on FT3D.

\begin{table}[t!]
\centering{
\resizebox{\columnwidth}{!}{
\begin{tabular}{|l|l|l|l|}
\hline
    \textbf{Ablation / Dataset} & {DAVIS2016} & {FBMS59} & {SegTrackV2} \\ \hline \hline
    Full Model LT-MS-K4 & 72.4 & 58.2 & 60.4 \\ \hline
    Unet3D only & 71.3 & 55.5 & 57.3 \\ \hline
    No consistency term $\mathcal{L}_c$ & 71.0 & 55.5 & 57.5 \\ \hline
    Polynomial space-time model & 69.8 & 54.5 & 56.6 \\ \hline
\end{tabular}
}
}
\vspace{0.15cm}
\caption{Ablation study for three main components of our method LT-MS ($K=4$) on DAVIS2016, FBMS59 and SegTrackV2. Only one model component is modified at a time. Input sequence length is 120 (in practice, it comprises the whole video for the DAVIS2016 dataset). The metrics used is the Jaccard index $\mathcal{J}$. All evaluations are performed on the binary ground truth for the three datasets as explained in Section \ref{stability}.
 }
 \label{tab:ablation}
\end{table}

\section{Experimental Results}
\label{results}

Our LT-MS method is above all a long-term multiple motion segmentation method. However, in order to quantitatively assess its performance and in absence of OFS benchmarks, we need to resort to VOS benchmarks.
We have carried out several types of experiments on four datasets: DAVIS2016\footnote{https://davischallenge.org/index.html} \cite{perazzi_benchmark_2016}, SegTrackV2\footnote{https://paperswithcode.com/dataset/segtrack-v2-1} \cite{li2013}, FBMS59 \cite{ochs2014}, and DAVIS2017-motion \cite{xie2022}.
However, most VOS benchmarks are concerned with the segmentation of the primary moving object in the foreground, which means a binary ground truth. This led us first to test our LT-MS method in a two-mask ($K=2$) configuration, as discussed in Section \ref{evaluation}, and reported in Table \ref{tab:results}. Then, we have performed multiple motion segmentation evaluation, especially on DAVIS2017-motion, as commented in Section \ref{multi-segment} and Table \ref{res:multi-segmentation}. Finally, we have assessed the long-term impact of LT-MS both on a quantitative (Table \ref{tab:stability}, Section \ref{stability}) and qualitative (Section \ref{qualitative-eval}) basis.

\begin{figure*}[tbh!]
\centering
\includegraphics[width=0.80\linewidth]{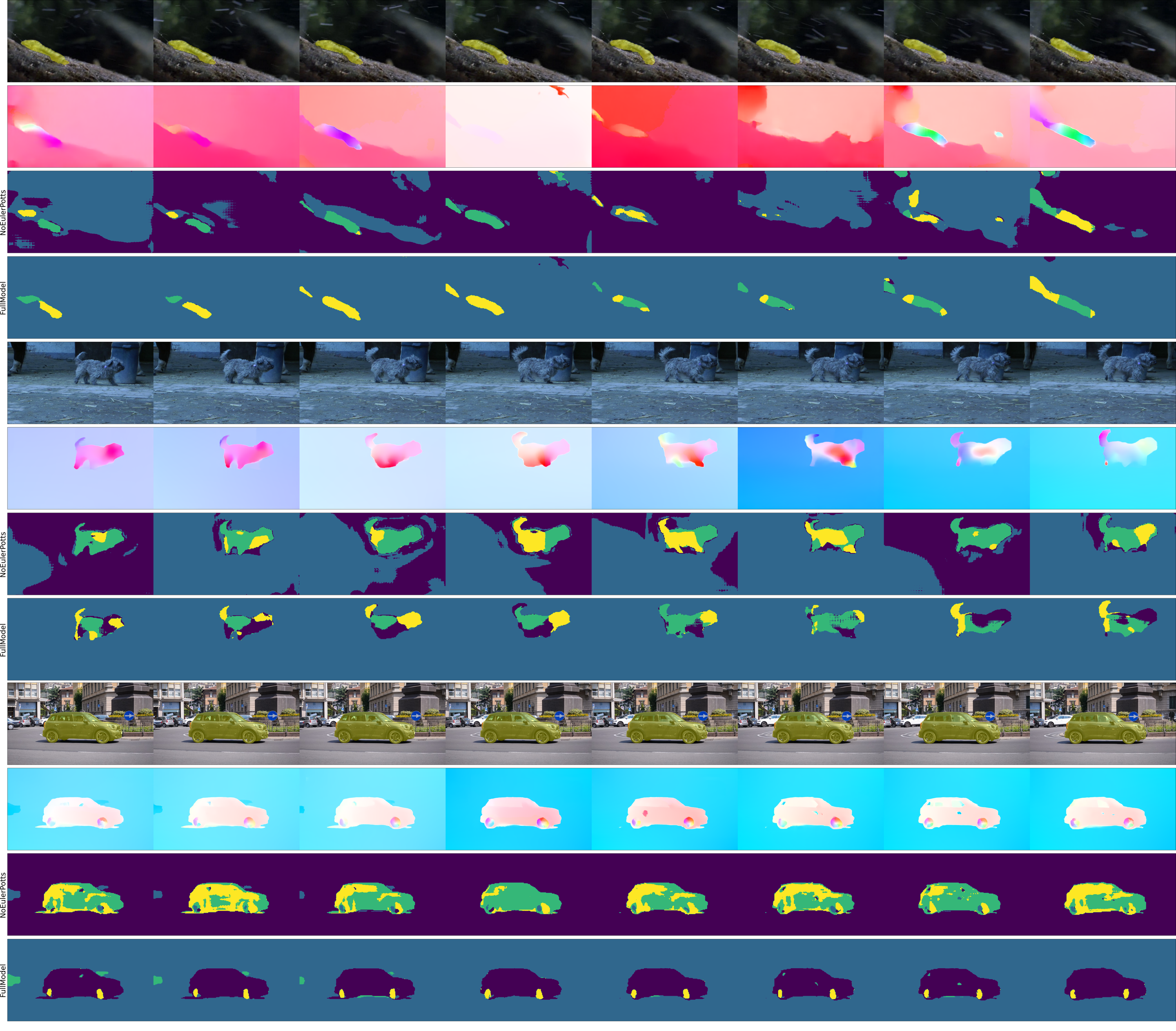}
\caption{Three groups of qualitative results regarding the ablation of the temporal-consistency loss term ($K=4$). They respectively correspond to the \textit{worm} video of SegTrackV2, the \textit{dogs01} video of FBMS59, and the %
\textit{car-roundabout} videos of DAVIS2016. For each group, the first row contains sample images with the segmentation ground-truth
overlaid in yellow, the second row displays the input flows, the third and fourth rows show the predicted motion segmentations, respectively without and with the temporal consistency loss term. Clearly, this model component allows us to get far more consistent segments over time.}
\label{tc-ablation}
\end{figure*}

\subsection{Datasets}
\label{dataset}
DAVIS2016 comprises 50 videos (for a total of 3455 frames), depicting diverse moving objects. It is split in a training set of 30 videos and a validation set of 20 videos. Only the primary moving object is annotated in the ground truth. The criteria for evaluation on this dataset are the Jaccard score $\mathcal{J}$ (intersection-over-union), and the contour accuracy score $\mathcal{F}$, the higher, the better for both.
FBMS59 includes 59 videos (720 annotated frames), and SegTrackV2 14 videos (1066 annotated frames). Both mostly involve one foreground moving object, but sometimes a couple of moving objects. For FBMS59, we use the 30 sequences of the validation set. Although annotations may comprise multiple objects, the VOS community exploits SegTrackV2 similarily as DAVIS2016, considering a binary ground-truth by grouping moving objects in the foreground. We follow this practice. However, we provide a multi-mask evaluation for FBMS59 as other works did.

DAVIS2017 is an extension of DAVIS2016, containing 90 videos. It includes additional videos with multiple moving objects, and it provides multiple-segment annotations for the ground truth. It is split into 60 videos for training and 30 for evaluation. DAVIS2017-motion is a curated version of the DAVIS2017 dataset proposed in \cite{xie2022} for a fair evaluation of methods based on optical flow only. Connected objects exhibiting the same motion are merged in the ground truth for evaluation. We proceed to the evaluation
with the official algorithm that involves a Hungarian matching process.

\subsection{Ablation study}
We have conducted an ablation study to assess three main components of our long-term multiple-motion segmentation method LT-MS. To this end, we took four masks ($K=4$), and call LT-MS-K4 this version of our LT-MS method. These three components are mainly related to the temporal dimension of the problem. We have changed one component at a time as specified hereafter.
We use the polynomial space-time quadratic motion model of ST-MS \cite{meunier2023-cvpr} instead of the space-time motion model based on B-splines over the input sequence.
We omit the consistency term $\mathcal{L}_c$ in the loss function.
We just take the convnet without the transformer decoder.
All the ablation experiments were run on the three datasets, DAVIS2016, FBMS59, SegTrackV2. Results are collected in Table \ref{tab:ablation}. We performed them for the input sequence length of 120 (in practice, the whole video for the DAVIS2016 dataset).

We can observe that the three ablations have almost the same impact on the performance. The three corresponding model components, i.e., the spline-based motion model, the temporal-consistency loss term, and the transformer decoder are thus all beneficial in similar proportions. They are able to handle the temporal dimension of the problem and the temporal motion evolution along the sequence in a compelling way. 
The contributions of these three components are even more significant for the FBMS59 and the SegTrackV2 datasets. In fact, the dynamic content of the majority of DAVIS2016 videos, along with performance scores calculated on a frame basis on these videos, do not allow to fully appreciate the contributions of the three model components.
Yet, they can be highlighted by visualizing results on some videos of DAVIS2016, as shown
in Fig.\ref{tc-ablation}.

\begin{table*}[tb!]
\resizebox{\columnwidth}{!}{
\centering{
    \begin{tabular}{|l || l|l|c|c|}
    \hline
        Methods & Training & Input & DAVIS2016 & SegTrackV2  \Tstrut \\
        ~ & ~ & ~ & $\mathcal{J}$ $\uparrow$ &   $\mathcal{J}$ $\uparrow$  \Bstrut\\ \hline
        {\bf Ours LT-MS-K2} & Unsupervised & Flow & 70.7 & 58.9 \Tstrut\Bstrut\\ \cline{4-5}
        EM \cite{meunier2022} & & & 69.3 & 55.5   \Tstrut\Bstrut\\ \cline{4-5}
        MoSeg \cite{yang_motion-grouping_2021} & &  & 68.3 & 58.6  \Tstrut\Bstrut\\  \cline{4-5}
        FTS \cite{papazoglou_fast_2013} & &  & 55.8 & 47.8   \Tstrut\Bstrut\\ \cline{4-5}
        TIS$_0$ \cite{griffin_tukey-inspired_2019} & &  & 56.2 & -  \Tstrut\Bstrut\\ \cline{1-2} \cline{4-5}
        OCLR \cite{xie2022} (flow only) & Additional Supervision &  & 72.1 & 67.6 \Tstrut\Bstrut\\ \cline{4-5}
         DyStab - Dyn \cite{dystab2021} & & & 62.4 & 40.0  \Tstrut\Bstrut\\ \cline{4-5}
         \cline{1-1} \cline{2-5}
        \hline \hline
         GWM \cite{choudhury2022} & Unsupervised & {RGB (Flow in loss)} & 79.5 & 78.3 \Tstrut\Bstrut\\ \cline{4-5}
        RCF \cite{lian2023} & &  & 80.9 & 76.7   \Tstrut\Bstrut\\ \cline{4-5}
        AMD \cite{liu2021} & &  & 57.8 & 57.0  \Tstrut\Bstrut\\ \cline{4-5}
        MOD \cite{ding2022} & &  & 73.9 & 62.2 \Tstrut\Bstrut\\ \cline{1-1} \cline{3-5}
         DivA(4)$\dagger$ \cite{lao2023} & & {RGB \& Flow}& 72.4 & 64.6 \Tstrut\Bstrut \\ \cline{4-5}
        IKE \cite{haller2021} & & & 76.0 & 66.4 \Tstrut\Bstrut \\ \cline{4-5} 
        TIS$_s$ \cite{griffin_tukey-inspired_2019} & &   & 62.6 & -   \Tstrut\Bstrut\\  \cline{4-5}
        CIS - No Post \cite{yang_unsupervised_2019} & & & 59.2 & 45.6   \Tstrut\Bstrut\\ \cline{1-2} \cline{4-5}
        DS  \cite{ye2022} & Optimisation at test time  & & 79.1 & 72.1  \Tstrut\Bstrut \\ \cline{1-2} \cline{4-5}
        GO \cite{ponimatkin2023}  & Optimisation at test time \& Additional Supervision & & 80.2 & 74.9  \Tstrut\Bstrut \\ \cline{4-5}
        \hline %
    \end{tabular}
   }
    }
\vspace{0.15cm}
\caption{Results obtained with our LT-MS-K2 method (with two masks) on DAVIS2016 and SegTrackV2.
An input flow sequence length of 120 is considered.
We include comparison with different categories of methods (scores from cited articles). The most relevant comparison is with the category of unsupervised methods taking only optical flow as input. All scores
correspond to evaluation on the binary ground-truth.
The Jaccard index $\mathcal{J}$ is the intersection over union between the extracted segments and the ground truth, the higher the better.
 Performance is assessed by the average score over all samples for SegTrackV2. For DAVIS2016, the overall score is given by the average of sequence scores.
$\dagger$DivA somehow uses RGB input since its conditional decoder leverages input images.}
 \label{tab:results}
 \end{table*}

\subsection{Quantitative and comparative binary evaluation}
\label{evaluation}
We first report in Table \ref{tab:results} the quantitative evaluation on DAVIS2016 and SegTrackV2, performed on the binary ground truth (foreground moving object \textit{vs} background) for the two datasets. This is the most frequent evaluation reported in the VOS literature. Therefore, we first apply our LT-MS method with two masks ($K=2$), denoted LT-MS-K2. Since our OFS does not establish the assignment of the two segments to respectively the background and the primary moving object, we rely on a simple heuristic designating the biggest segment as the background one.

Let us recall that our actual objective is the optical flow segmentation (OFS) task, even if we evaluate our method on VOS benchmarks in absence of OFS benchmarks. Since the two VOS benchmarks deal with the segmentation of the primary object moving in the foreground, it may occur discrepancies with OFS, which negatively impacts the scores for OFS methods. If wrong w.r.t. VOS ground truth, the segmentation of additional moving parts actually makes sense from the OFS standpoint, such as moving objects in the background, subparts of moving articulated bodies, or static foreground objects exhibiting motion parallax with a moving camera. Therefore, the comparative evaluation regarding OFS must be restricted to unsupervised methods taking only optical flow as input, namely, the top part of Table \ref{tab:results}. Nevertheless, for a more comprehensive view, we also include results of VOS methods that leverage both RGB images and flow as input, in the bottom part of Table \ref{tab:results}.

We evaluate LT-MS-K2 on a sequence basis, not on a frame basis as done by the other methods. This is more demanding, but we have done so to be consistent with the long-term intrinsic property of our method. We took a sequence length $T=120$ for DAVIS2016 and SegTrackV2.
Table \ref{tab:results} also collects results obtained by other existing unsupervised methods also with two masks, when available on these benchmarks.
We keep the categorization that we proposed in \cite{meunier2022} regarding input and training. Nevertheless, we have added a category w.r.t. the network input for four recent methods, \cite{choudhury2022,ding2022,lian2023,liu2021}, that only use RGB images as imput at test time, the optical flow being only involved in the loss function. 
OCLR leverages additional supervision, since it relies on human-annotated sprites to include realistic shapes in the computer-generated data used at training time. 
We consider the OCLR version taking only optical flow as input.
The same holds for the DyStab method, the DyStab-Dyn version is considered.
The post-processing added to the CIS method \cite{yang_unsupervised_2019}, based on Conditional Random Fields (CRF), is an heavy one, which has led most authors to retain only the version without post-processing for a fair comparison.
We also include a separate category for two recent unsupervised methods, DS \cite{ye2022} and GO \cite{ponimatkin2023},
where optimisation is performed at test time.

As shown in Table \ref{tab:results}, LT-MS-K2 provides the best results on the two datasets
in the category of unsupervised methods based on optical flow only.
Besides, we report a repeatability study in the Appendix.
OCLR and DivA demonstrate better performance, especially on the SegTrackV2 dataset. However, as aforementioned, OCLR involves additional supervision, while DivA leverages RGB images in its conditional decoder. In addition, DivA, along with MoSeg and CIS methods, takes multi-step flows as input between $t$ and in turn $t+1$, $t+2$, $t-1$, $t-2$, and averages the four corresponding predictions to get the final result.

Failure cases may occur for very complex motion (e.g., ripples on the water), or a particularly strong temporal evolution (as an object approaching the camera from afar quickly and closely). 
Overall, temporal consistency is properly handled over long periods by our LT-MS method.
We want to stress that our method is the only one providing \textit{by design} a coherent segmentation over the sequence, which is a significant added value for downstream tasks, like tracking or dynamic scene interpretation.

\begin{table}[!t]
    \centering{
    \resizebox{\columnwidth}{!}{
    \begin{tabular}{|l||c c c|c c|}
    \hline
        Methods  & \multicolumn{3}{c|} {DAVIS2017-motion} & \multicolumn{2}{c|} {FBMS59} \\ 
         & $\mathcal{J}\&\mathcal{F}$ $\uparrow$ & $\mathcal{J}$ $\uparrow$ & $\mathcal{F}$ $\uparrow$ & bIoU & Linear Assign. \\ \hline \hline
        {\bf Ours LT-MS-K3} & 42.2 & 39.3 & 45.0 & 58.4 & 47.2 \\ \hline
        ST-MS \cite{meunier2023-cvpr} & 42.0 & 38.8 & 45.2 & - & - \\ \hline
        MoSeg \cite{yang_motion-grouping_2021} & 35.8 & 38.4 & 33.2 & - & - \\ \hline
        OCLR \cite{xie2022} & 55.1 & 54.5 & 55.7 & - & - \\ \hline
        DivA \cite{lao2023} & - & - & - & 42.0 & -    \\\hline  
    \end{tabular}
    }}
    \vspace{0.15cm}
    \caption{Multi-segment evaluation, all compared methods using three masks.
Evaluation is performed on the video as a whole following the official DAVIS2017 scheme. Reported scores are the average of the individual video scores. The $\mathcal{F}$ metrics focuses on segment boundary accuracy. $\mathcal{J}\&\mathcal{F}$ is the mean of the two. 
For FBMS59, the evaluation metrics are bootstrap IoU (bIoU) \cite{lao2023} where each ground-truth mask is mapped to the most likely predicted segment (performed at the frame level), and linear assignment that is a bilinear mapping between the ground-truth and the predicted segments at the sequence level (as in the official DAVIS2017 evaluation).
}
\label{res:multi-segmentation}
\end{table}

\subsection{Multi-segment evaluation}
\label{multi-segment}
We now focus on the multiple motion segmentation for the DAVIS2017-motion and FBMS59 datasets.
Since multiple-motion segmentation is harder than the binary motion segmentation that only involves moving foreground \textit{vs} background, accuracy scores are expected to decrease for all methods. In Table %
\ref{res:multi-segmentation}, we report comparative results on the DAVIS2017-motion dataset using the official evaluation procedure, and on FBMS59. As for the other methods, we performed the multiple motion segmentation with three masks ($K=3$), and we name this version LT-MS-K3. To this end, we finetuned the LT-MS-K4 network on the DAVIS2016 training set with now three masks ($K=3$). The resulting performance on DAVIS2017-motion is slightly better than ST-MS and far better than MoSeg. Let us recall that OCLR involves additional supervision. Regarding FBMS59,
we report multimask segmentation results with two different metrics. %
Our LT-MS method outperforms by a large margin DivA \cite{lao2023} that also attempted this multi-segment evaluation.

\begin{figure*}[t!]
\centering{
\includegraphics[width=0.80\linewidth]{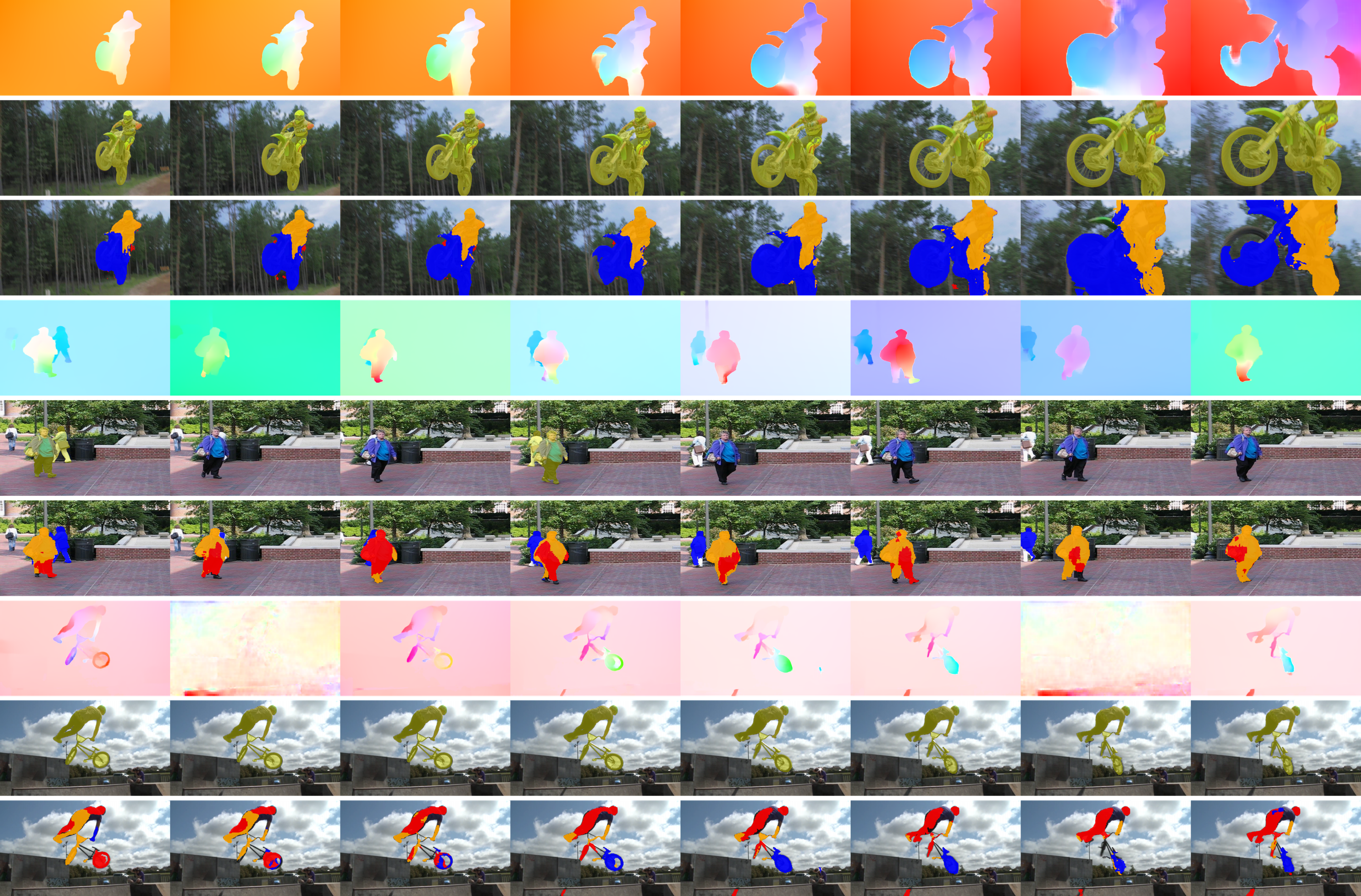}
\caption{\footnotesize{Results obtained with our LT-MS-K4 method %
for \textit{motocross-jump} from DAVIS2016, \textit{people02} from FBMS59 and \textit{bmx} from SegTrackV2. For each group, the first row samples flow fields (HSV color code) corresponding to the processed video. The second row contains the corresponding images of the video, where ground-truth of the moving object is overlaid in yellow.
The third row shows the motion segments provided by our LT-MS-K4 method
with one colour per segment.
We constantly adopt the same color set for the three masks corresponding to moving objects (blue, red and orange), and we let the image for the background mask.
}}
\label{fig:4masks}
}
\end{figure*}

\begin{table}[bt!]
\centering{
\begin{tabular}{|l||c c|c c|c c|}
\hline
    \textbf{Methods} & \multicolumn{2}{c|}{DAVIS2016} & \multicolumn{2}{c|}{FBMS59} & \multicolumn{2}{c|}{SegTrackV2} \\ \hline
    \textbf{Evaluation basis} & f & s & f & s & f & s \\ \hline \hline
    {LT-MS-K4} & 75.0 & 72.4 & 60.1 & 58.2 & 63.4 & 60.4\\ \hline
    {ST-MS} \cite{meunier2023-cvpr} & 71.4 & 70.2 & 56.3 & 55.2 & 55.3 & 53.7 \\ \hline
    {EM} \cite{meunier2022} & 77.1 & 68.0 & 66.1 & 51.9 & 62.5 & 50.6 \\ \hline
    \end{tabular}
}
\vspace{0.15cm}
\caption{Accuracy and stability in the evaluation, on DAVIS2016, FBMS59 and SegTrackV2, of our three multiple motion segmentation models, LT-MS, ST-MS and EM, all with $K=4$, on a frame-based (f) basis and on a sequence-based (s) basis for a sequence length of 120. Reported scores correspond to the $\mathcal{J}$ metrics.
}
\label{tab:stability}
\end{table}

\subsection{Accuracy \textit{vs} stability}
\label{stability}
Since the frame-based quantitative evaluation reported in Section \ref{evaluation} cannot fully reflect the contributions of our long-term motion segmentation model, we have carried out complementary experiments. The idea is to somehow cross-validate accuracy and stability. To this end, we compare our three models, the frame-based EM model, the short-term ST-MS model, and the long-term LT-MS model. We proceed in two ways. First, we evaluate all three models on a frame-based basis,
then, on a sequence-based basis for a sequence length of 120. By sequence length, we mean the length of the flow sequence taken as input for LT-MS or the output length of the temporal linkage (post-processing stage) for ST-MS.

Results obtained on DAVIS2016, FBMS59 and SegTrackV2 are shown in Table \ref{tab:stability} for a binary ground truth. In the Appendix, we explain how we select the segments for the binary evaluation from the set of segments delivered by our multiple motion segmentation methods, when using a number of masks $K > 2$. It is not the same procedure as the one used at the end of Section \ref{evaluation}, which explains the scores obtained for LT-MS-K4 compared to Table \ref{tab:results}. Here, we can select the right segments by simply comparing with the ground-truth segments, since it is a question of internal comparison. As expected, the frame-based evaluation favours a frame-based multiple motion segmentation, and overall, the accuracy slightly decreases with the sequence length since it is more demanding to maintain the same labelling, and then to preserve accuracy, over a sequence. In contrast, the sequence-based evaluation highlights the temporal stability brought by ST-MS and with a significant larger margin by LT-MS,
in contrast to EM whose performance significantly drops when evaluated on a sequence basis.

\subsection{Qualitative visual evaluation}
\label{qualitative-eval}
Visual results are provided for LT-MS-K2 in the Appendix. Fig.\ref{fig:4masks} contains a set of visual results to demonstrate how our LT-MS-K4 method behaves on different situations. We display three result samples obtained on different videos of the benchmarks.
We can observe that the motion segments are globally accurate.
Since our method involves several masks, we can properly handle articulated motions (\textit{people02, bmx}), deal with the presence of several moving objects in the scene (\textit{people02}), separate the rider from the bike or motorbike (\textit{bmx, motocross-jump}).
Since our method enforces temporal consistency, it can deal with errors in optical flow estimation or with objects that momentarily stop moving (\textit{bmx}). 

\begin{figure*}[t!]
\centering{
\includegraphics[width=0.80\linewidth]{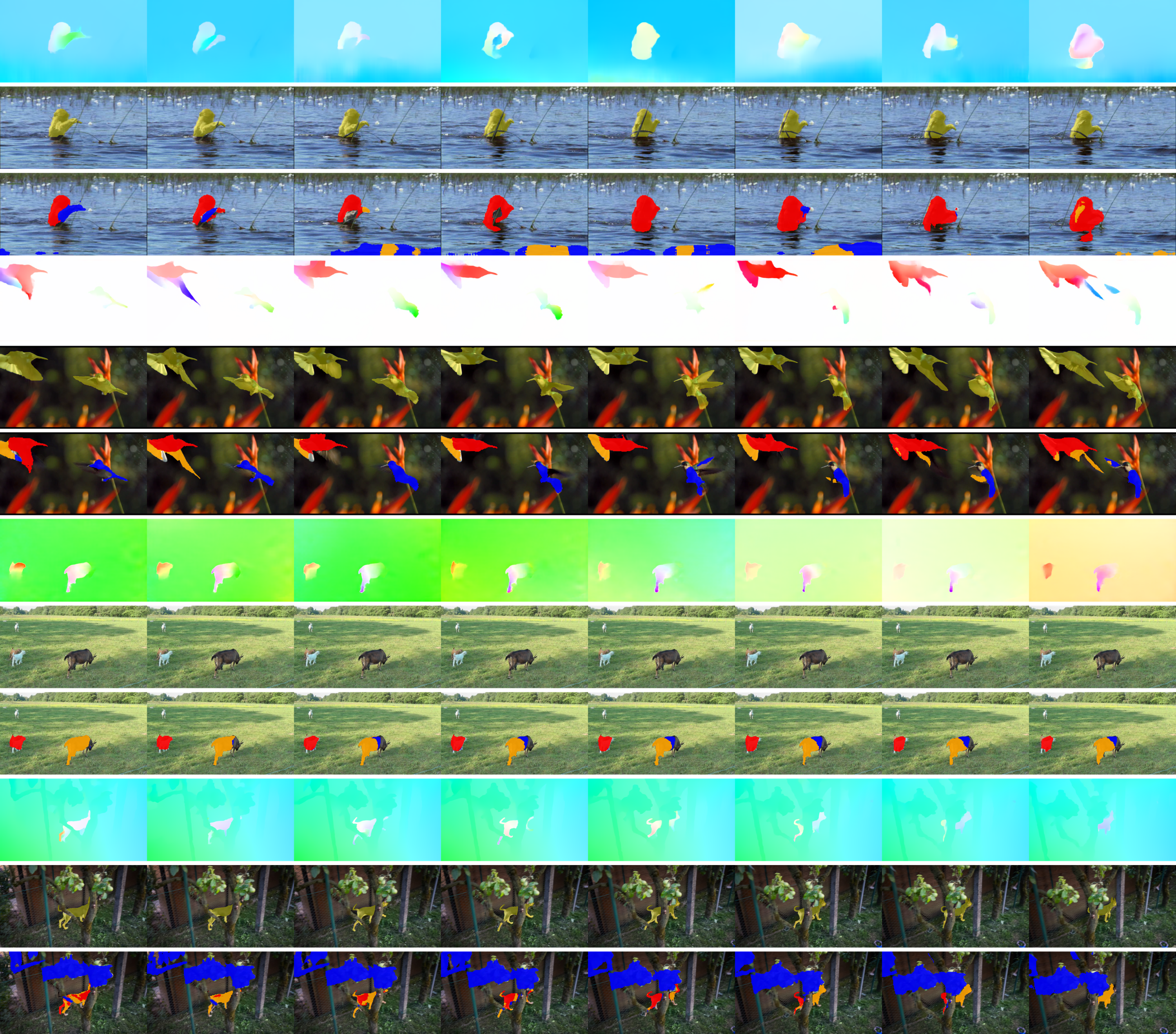}
\caption{\footnotesize{Results obtained with our LT-MS-K4 method ($K=4$). Four groups of results are displayed: \textit{monkey}, \textit{hummingbird} from SegTrackV2, \textit{goats01} from FBMS59 and \textit{libby} from DAVIS2016. For each group, the first row samples successive flow fields (HSV color code) corresponding to the processed video. The second row contains the corresponding images of the video, where the ground-truth of the moving object is overlaid in yellow (when available at that frame). The third row shows the motion segments provided by our LT-MS-K4 method
with one colour per segment. For all the results, we adopt the same color set for the three masks corresponding to the moving objects (blue, red and orange), and we let the background image for the background mask.
}}
\label{fig:4masks-add}
}
\end{figure*}

\begin{figure*}[h!]
\centering{
\includegraphics[width=0.80\linewidth]{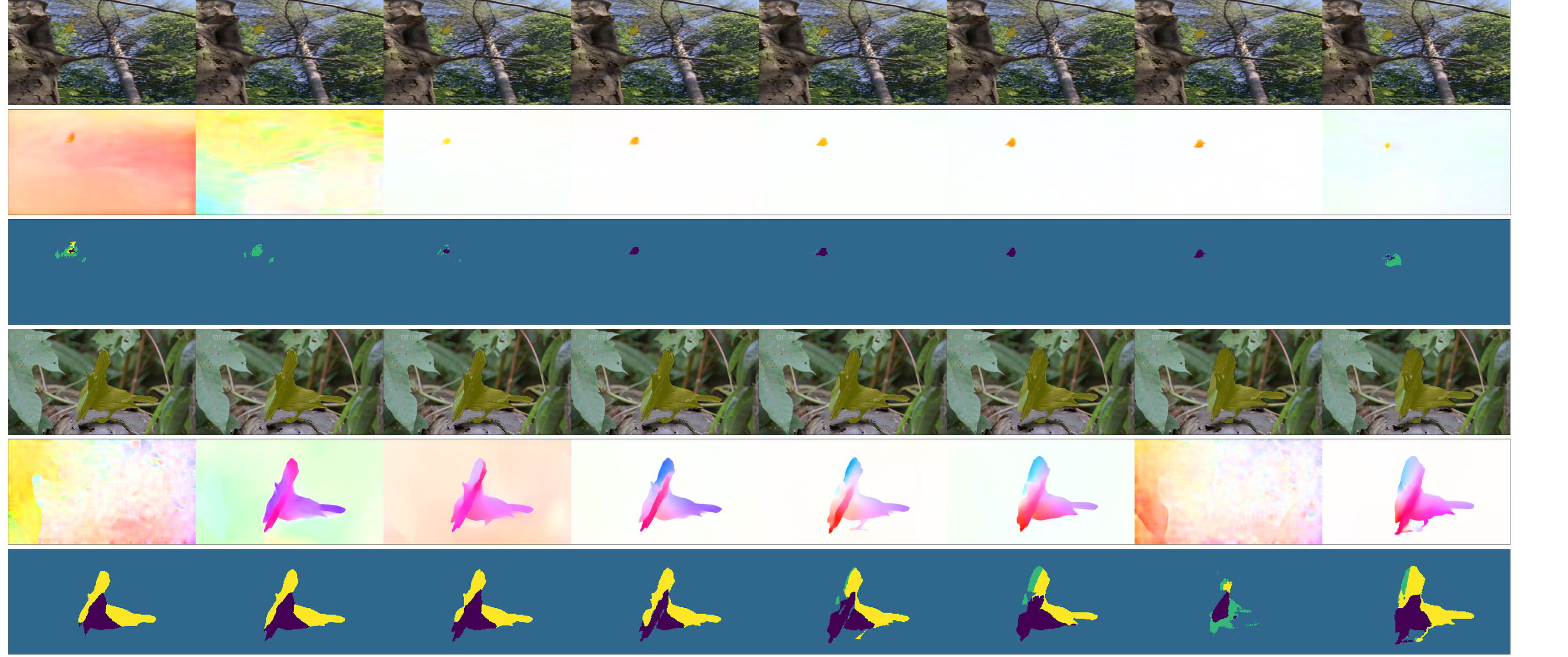}
\caption{Illustration of the temporal consistency provided by our LT-MS-K4 method on two videos of SegTrackV2, \textit{birdfall}, \textit{bird-of-paradise},
For each group, the first row contains the video images with the ground truth overlaid in yellow;
the second row depicts the corresponding flow fields represented with the HSV code while normalized independently from each other; the third row provides the predicted segmentation.}
\label{cases-SegTrackV2}
}
\end{figure*}

Additional visual results of our LT-MS-K4 are collected in Fig.\ref{fig:4masks-add}.
Again, we can properly cope with articulated motions (\textit{monkey}), with the presence of several moving objects in the scene (\textit{hummingbird}, \textit{goats01}), or accomodate motion parallax (\textit{libby}).
Finally, in Fig.\ref{cases-SegTrackV2}, we collect visual results on two videos of the SegTrackV2 dataset, from top to bottom, \textit{birdfall} and \textit{bird-of-paradise} videos.
Results demontrate the ability of our long-term segmentation method to recover correct segmentation even in case of flow fields wrongly estimated for some frames. By the way, this ability was also demonstrated by the greater accuracy (Tables \ref{tab:results} and \ref{tab:stability}) obtained on SegTrackV2 dataset by LT-MS compared to EM that operates on a frame basis.

\section{Conclusion}
\label{conclusion}

We have designed an original transformer-based unsupervised method for segmenting multiple motions over a sequence, without any postprocessing. It performs it in one go, which is a more demanding goal, but provides decisive added value for downstream tasks like tracking or dynamic scene interpretation. Our LT-MS method leverages the ELBO framework for the loss function, and fully acknowledges the temporal dimension of the  motion segmentation problem.
Our method
delivers a stable and consistent OF segmentation throughout long video sequences. It introduces at training time B-splines spatio-temporal parametric motion models over space-time volumes, and a loss term expressing temporal consistency over successive masks while taking care of occlusions.
Our transformer-based network
can be applied at test time  to input volumes of any time length.
It can accomodate different choices on the number $K$ of masks with a simple fine-tuning step.
Experimental results on several datasets demonstrate the efficiency and the accuracy of our LT-MS method by providing competitive results on several datasets. In addition, it is very fast at test time.

\section{Appendix}
\subsection{Loss function}

\subsubsection{Experimental validation of the approximation of the temporal consistency loss term.}

We have tried to assess the impact of the approximation made in the definition of the temporal consistency term of the loss function as developed in Section 3.2 of the main text. To this end, we have compared the full LT-MS-K4 model with a version where the temporal consistency term is issued from eq.(11) in the main text before the approximation. More specifically, we consider the alternate loss function given by:

\begin{align}
    &\mathcal{L}(f) = \sum_{t=1}^T \sum_{i\in \Omega} \sum_{k=1}^K g_{\phi}(f)_{i,t}^k ||f(i,t) - \tilde{f}_{S_n(\theta_k)}(i,t)||_1 / \xi_{f, t} \nonumber \\
    &-\gamma_1 \sum_{i\in \Omega}\sum_{t=2}^T\sum_{k=1}^K g_{i,t}^kg_{i,t-1}^k  + \gamma_2 \sum_{t=1}^T \sum_{i\in \Omega} \sum_{k=1}^K g_{\phi}(f)_{i,t}^k \log g_{\phi}(f)_{i,t}^k,
\label{wo-approxim}
\end{align}
where $\xi_{f,t} = \sum_{i\in\Omega} ||f(i,t)||_1$.
In practice, we set $\gamma_1 = 0.5$ and $\gamma_2 = 0.01$.

We carried out the comparison with an input flow sequence length of 120 on th ethree datasets DAVIS2016, FBMS59 and SegTrackV2. We obtained the results reported in Table \ref{tab:ablation-add}. Results are almost on par for DAVIS2016. The increased performance of the full model LT-MS-K4 is slightly higher for the other two datasets, most likely because the L1 norm used in our loss is more robust to error noise in the network predictions during training.

\begin{table}[b!]
\label{tab:ablation-add}
\centering{
\begin{tabular}{|l|l|l|l|l|}
\hline
    \textbf{Prior / Dataset} & {DAVIS2016} & {FBMS59} & {SegTrackV2} \\ \hline
    Full Model LT-MS-K4 & 72.4 & 58.2 & 60.4 \\ \hline
    Modified model of eq.(\ref{wo-approxim}) & 71.6 & 55.6 & 57.4 \\ \hline
\end{tabular}
}
\vspace{0.1cm}
\caption{Comparison of the two temporal consistency loss terms}
\end{table}

\subsection{Addition to the ablation study}
Fig.\ref{comp-motion-models} highlights the contribution of the spline-based motion model on the \textit{dog} video, and its obvious ability to handle motions that do not vary uniformly, as the erratic movement of the dog.

\subsection{Segment selection for binary evaluation}
\label{selection}
VOS and OFS are close but not identical tasks. However,
we use VOS benchmarks, since no OFS benchmarks are available. Let us recall that the VOS one is attached to the notion of a primary object of interest moving in the foreground (sometimes, a couple of objects). As a consequence, we have to select the right segments to cope with the binary ground truth of the VOS benchmarks, as usually done for the DAVIS2016, SegTrackV2 and FBMS59 datasets.

Since we deal with multiple-motion segmentation, i.e., $K$ segments with $k>2$, we have to group them into two clusters corresponding to the foreground moving object on one side and the background on the other side. We proceed as in the ST-MS method \cite{meunier2023-cvpr} on a sequence basis. Note that this selection is used solely for results presented in Table 4. of the main paper.

For the sequence, we select the subset $\mathcal{S^*}$ of labels (that is, of segments) to constitute the predicted foreground, using the binary ground-truth $g_t$. The selected label subset $\mathcal{S^*}$ is given by: 
\begin{align}
    \mathcal{S^*} = \arg \max_{\mathcal{S} \subset \mathcal{P}(\{1,..,K\})} \sum_{t=0}^T J(\bigcup_{l \in \mathcal{S}}l_t,g_t),
\end{align}
where $J$ is the IoU score between two segments,
$\mathcal{P}(\{1,..,K\})$ is the partition of the labels in the sequence, and the binary masks $l_t$ correspond to the label mask $m_t$. Once we have selected the subset $\mathcal{S^*}$ of labels, we can use it to build our binary segmentation $\{s_t, t =0, .., T\}$ on the whole sequence for evaluation.

Let us recall that for the case $K=2$,
we just pick the smallest of the two masks to compare it to the ground truth mask of the primary moving object.

\subsection{Multi-mask evaluation}

Regarding the evaluation on the DAVIS2017-motion dataset whose ground truth is not binary,
we use the official evaluation script (which performs Hungarian matching over the sequence) to associate each predicted segment $k$ with the ground-truth annotations, knowing that we have considered $K=3$ masks for this experimentation as the other methods we compare with. We also apply it to FBMS59 for the multiple segment setting.

\begin{figure}
\includegraphics[width=0.9\textwidth]{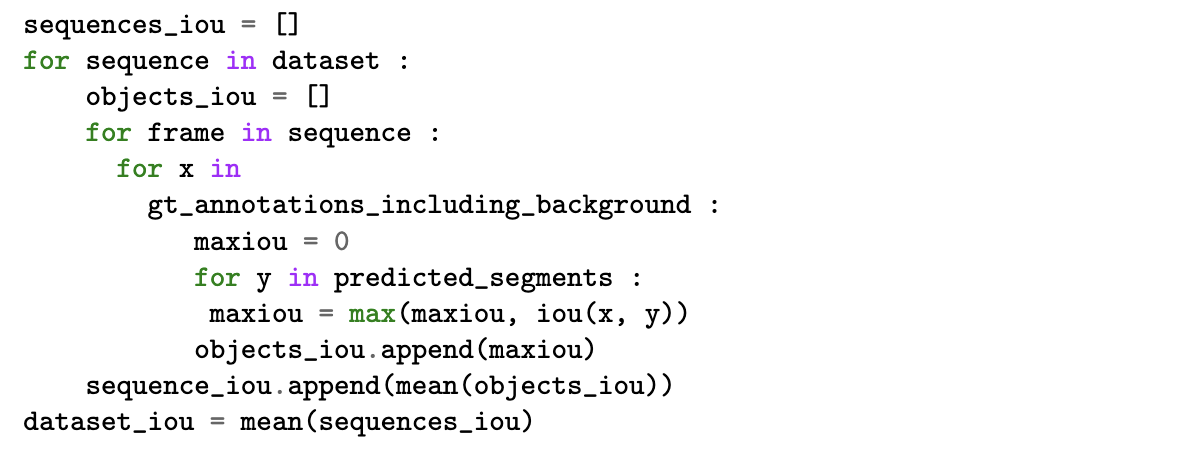}
\caption{Evaluation Boostrap IoU}
\label{fig:fbms_eval}

\end{figure}

In addition, we implement the "Boostrap IoU" score to evaluate multi-mask segmentation in FBMS59, as described in \cite{lao2023}\footnote{We use the multi-label ground truth provided by Dong Lao, first author of \cite{lao2023}.} The core idea is to match each ground-truth segment to the most likely segment (i.e., with the one of highest IoU). Let us note that this evaluation includes the background segment, since background is not identified in the FBMS59 multi-mask annotation.
The algorithm for evaluation we used is described in figure \ref{fig:fbms_eval}.

For the linear assignment score, we find the best bipartite sequence-level match between the ground truth and the predictions. Similar to the bIoU evaluation, we compute the score with all the labels of the ground truth. This evaluation is more demanding, because it forces a one-to-one match between the prediction and the masks at the sequence level.

\subsection{Qualitative results for LT-MS-K2}
\begin{figure*}[t!]
\centering{
\includegraphics[width=1.2\linewidth]{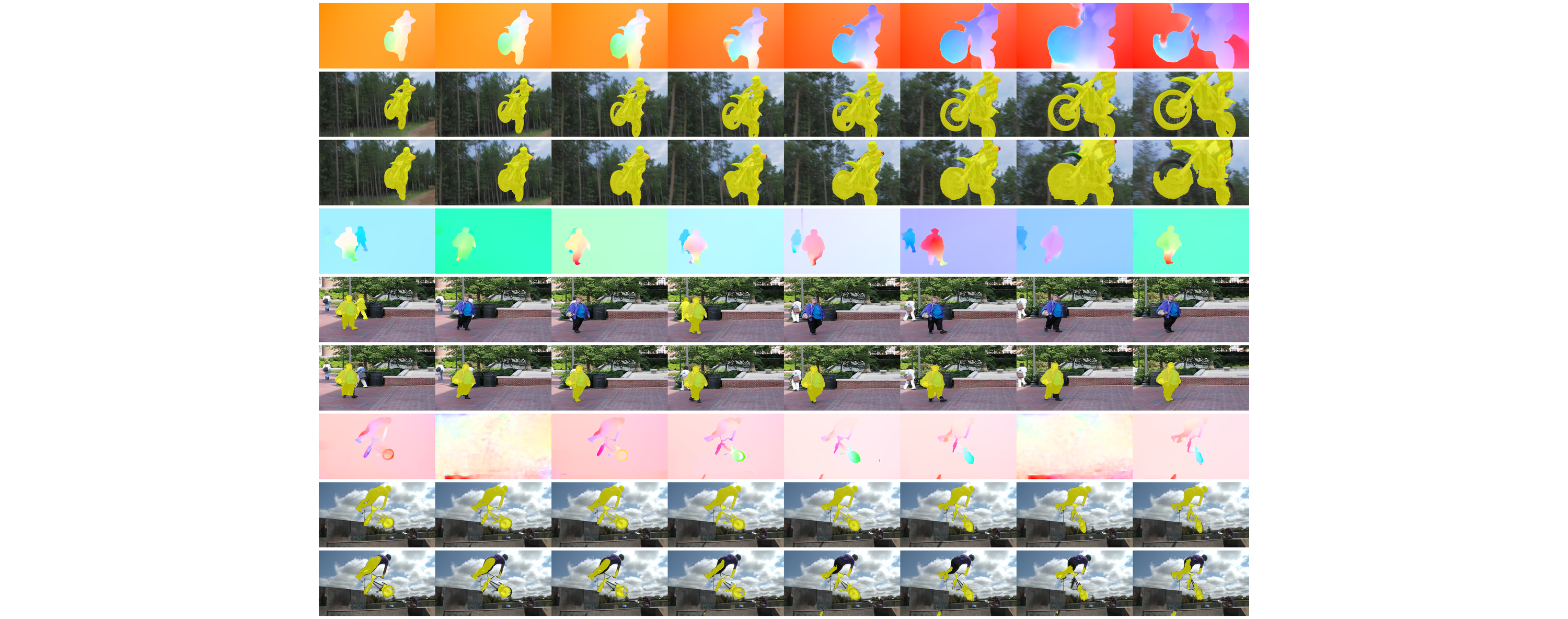}
\includegraphics[width=1.2\linewidth]{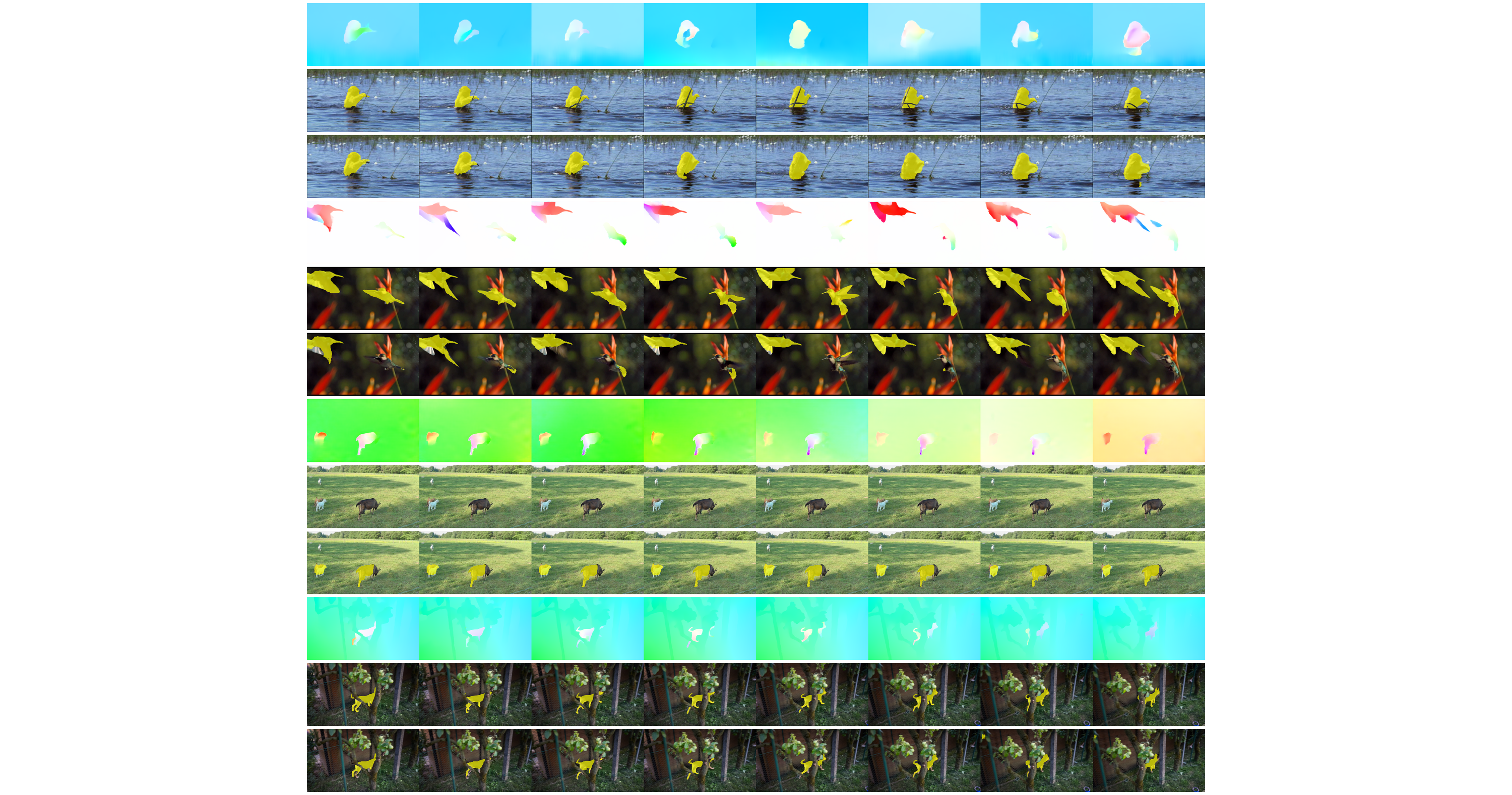}
\caption{
Results obtained with our LT-MS-K2 method %
for, from top to bottom, \textit{motocross-jump} from DAVIS2016, \textit{people02} from FBMS59, \textit{bmx} from SegTrackV2, \textit{monkey}, \textit{hummingbird} both from SegTrackV2, \textit{goats01} from FBMS59 and \textit{libby} from DAVIS2016. For each group, the first row samples flow fields (HSV color code) corresponding to the processed video. The second row contains the corresponding images of the video, where the binary ground-truth of the primary moving object is overlaid in yellow (when available, for FBMS59 the ground truth is not available in every frame).
The third row shows the binary segmentation provided by our LT-MS-K2 method.
We let the image content for the background mask.
}
\label{fig:2masks}
}
\end{figure*}

We report in Fig.\ref{fig:2masks} visual results obtained with our LT-MS-K2 method with two masks ($K=2$). LT-MS-K2 is able to accurately segment the primary moving object throughout the sequence for the videos \textit{motocross-jump, people2, monkey, hummingbird, goats01} and \textit{libby}. In \textit{bmx} video, it segments only part of the whole formed by the cyclist and his bike, but there are different motions within this articulated moving set, and one single motion model cannot handle all of them.
These results can be compared with those obtained by LT-MS-K4 on the same videos and reported in Figures 4 and 5 of the main text.

\subsection{Repeatability}

\begin{figure*}[h!]
    \centering{
    \includegraphics[width=0.9\textwidth]{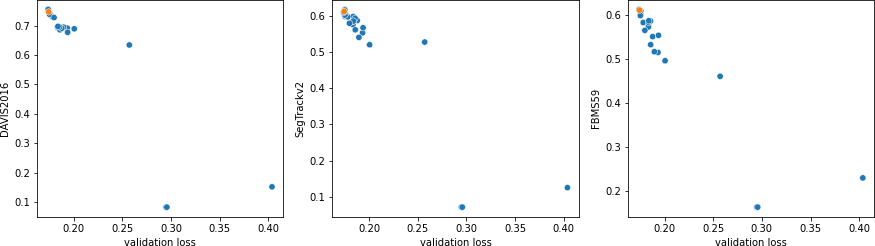}
    \includegraphics[width=0.9\textwidth]{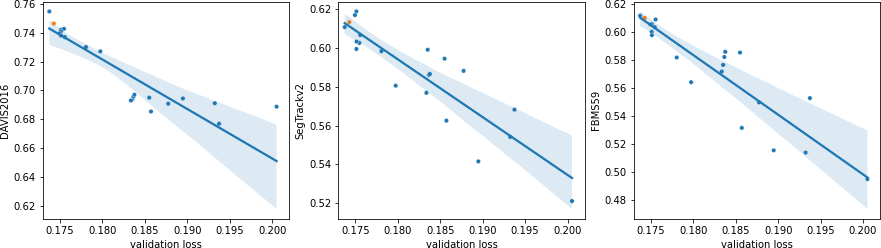}
    }
\caption{Model scores depending on initialization. Each dot represents a trained model. The $x$-axis represents the validation loss on the DAVIS2016 train dataset, and the $y$-axis the performance on the evaluation dataset. The top row includes all models and the bottom row excludes models that diverged during training (validation loss $> 0.25$). The bottom row displays the linear relationship between validation loss and evaluation score. The model whose results are reported in the main text is represented as an orange dot.}
\label{fig:reproducibility}
\end{figure*}

\begin{figure*}[h!]
    \centering
    \includegraphics[width=0.9\textwidth]{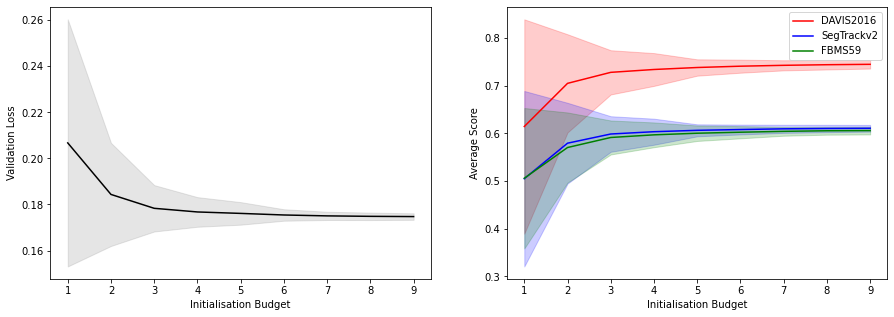}
    \caption{Evolution of the model performance as a function of the initialization budget. Left : the validation loss associated to the best network of each subset. Right : average performance on the test data set associated to this network for each different budget. The filled area correspond to +/- the standard deviation for each curve.}
    \label{fig:initialization-budget}
\end{figure*}

\begin{figure*}[p!]
\centering
\includegraphics[width=0.95\linewidth]{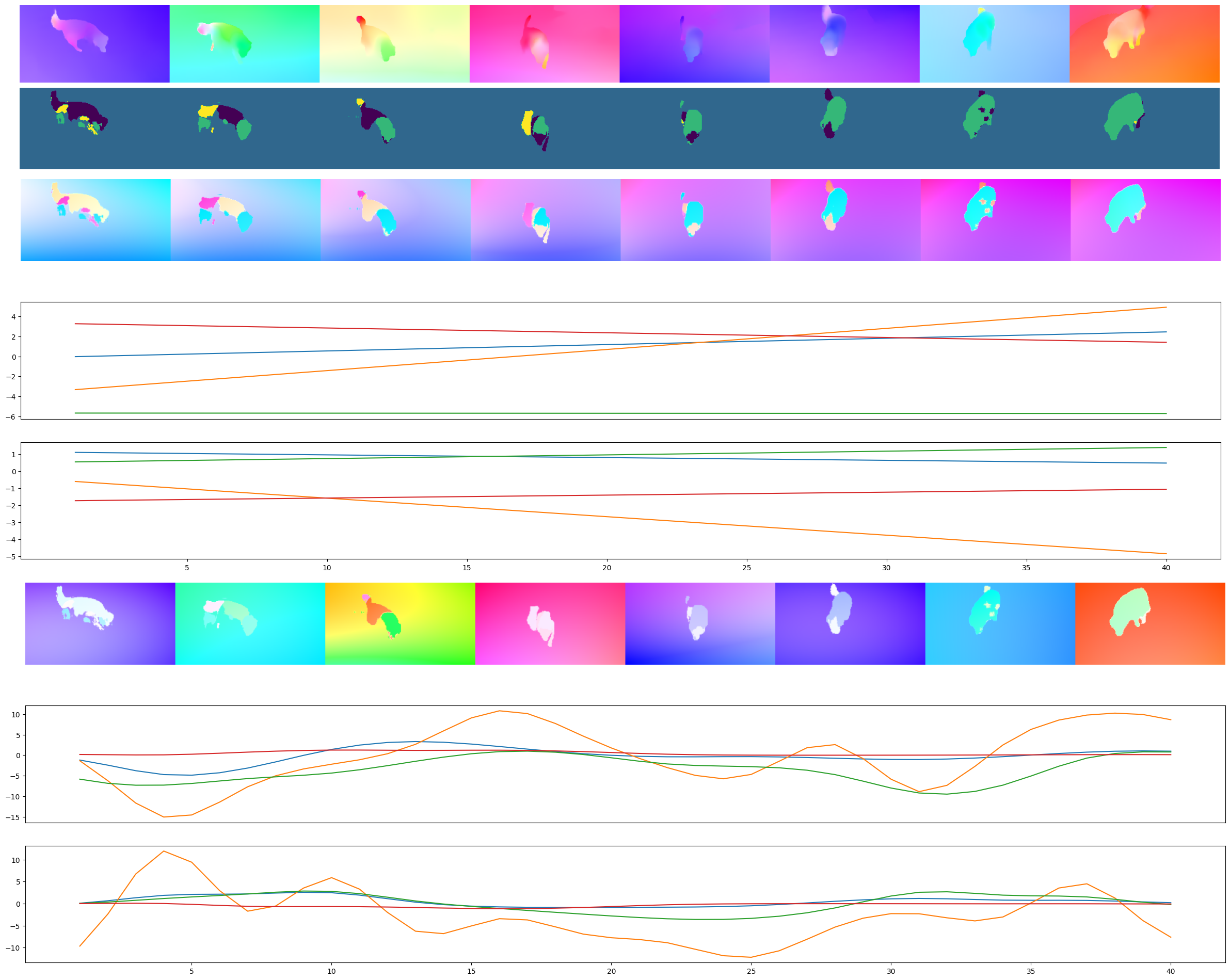}
\caption{Visual assessment of the contribution of the spline-based motion model on the \textit{dog} video of DAVIS2016 dataset (with $K=4$). From top to bottom: input flows of the sequence; corresponding predicted motion segmentation maps; flows reconstructed with the space-time polynomial motion model; plots over time of the mean-value of the $u$-components of the flows provided per segment by the space-time polynomial motion models; plots over time of the mean-value of the $v$-components of the flows provided per segment by the space-time polynomial motion models; flows reconstructed with the space-time spline-based motion model; plots over time of the mean-value of the $u$-components of the flows provided per segment by the spline-based motion models; plots over time of the mean-value of the $v$-components of the flows provided per segment by the spline-based motion models. Clearly, the space-time polynomial model fails to handle non-uniformly varying motion, whereas the spline-based model does.}
\label{comp-motion-models}
\end{figure*}

With the introduction of the transformer decoder, we experimentally found that the convergence of the network depends on the weights initialization, and that the same network and loss configuration can yield different results at test time. In Fig.\ref{fig:reproducibility}, we show that our unsupervised loss on a held-out validation set is a good indicator of the network performance at test time. This is a critical point for model and hyperparameter selection, since we do not have access to the ground truth at training time (with our fully unsupervised scenario), and thus we cannot evaluate model performance. Fig.\ref{fig:initialization-budget} plots the model performance after training as a function of the initialization budget.

In our evaluation experiments, we account for this randomness by training five models with the same set of seeds for all ablations and by reporting the score of the model with the lowest validation loss for each model.

\subsection{Training on real-world data}

\begin{table}[h!]
\label{tab:training}
\centering{
\begin{tabular}{|l|l|l|l|}
\hline
    \textbf{Training mode / Dataset} & {DAVIS2016} & {FBMS59} & {SegTrackV2} \\ \hline
LT-MS-K4 (FT3D) & 72.4 & 58.2 & 60.4 \\ \hline
LT-MS-K4 (DAVIS2016) & 72.8 & 57.6 & 61.3 \\ \hline
\end{tabular}
}
\vspace{0.1cm}
\caption{Training on synthetic (FT3D) and on real (DAVIS2016) data. Input sequence length of 120.}
\end{table}

We have trained our model LT-MS-K4 on the DAVIS2016 training set and evaluated it on DAVIS2016 validation set, FBMS59 and SegTrackV2.
Results are collected in Table \ref{tab:training} and compared with those obtained when training our model on the synthetic FT3D dataset. We can observe that performance when training on DAVIS2016 remains globally on par with performance when training on FT3D. Let us note that the volume of data is much smaller when training on DAVIS 2016.

\subsection{Influence of the input sequence length at inference}

\begin{figure*}[h!]
\centering
\includegraphics[width=\linewidth]{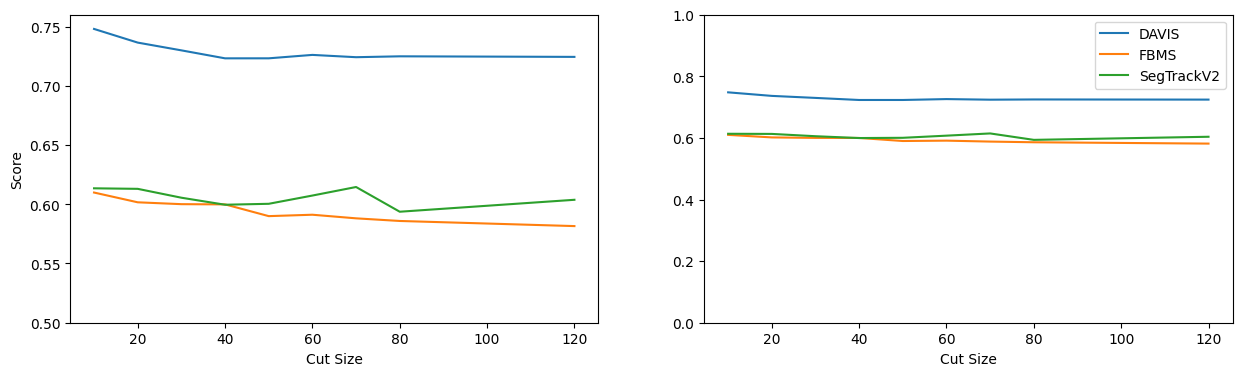}
\caption{Influence of the length of the input flow sequence (horizontal axis of the plot) on the LT-MS-K4 method for three datasets, DAVIS2016, FBMS59 and SegTrackV2. As expected, the best performance is obtained for the smallest length (equal to 10), but performance decreases slowly when the sequence length increases and remains stable for larger ones. On the right, the plot is shown with the full range of values on the ordinate $[0,100]$. On the left, a zoomed-in version of the plot.}
\label{CST-impact}
\end{figure*}

We have also tested how our LT-MS-K4 method behaves when varying the length of the input flow sequence. Results are plotted in Fig.\ref{CST-impact} for three datasets, DAVIS2016, FBMS59 and SegTrackV2.

As expected, the best performance is obtained for the smallest length (equal to 10) of the input flow sequence. However, performance decreases slowly when the length increases, and remains stable for larger ones. It demonstrates the intrinsic ability of the LT-MS method to achieve accurate and consistent motion segmentation over long periods of the video, which is a unique property.

\subsection{Detailed results for DAVIS2016}
\csvnames{davisnames}{1=\Sequence,2=\JM,3=\JO,4=\JD,5=\FM, 6=\FO, 7=\FD}
\csvstyle{Rta}{tabular=|l|c|c|c|c|c|c|,table head=\hline Video & $\mathcal{J}$ (M) & $\mathcal{J}$ (O) & $\mathcal{J}$ (D) & $\mathcal{F}$ (M) & $\mathcal{F}$ (O) & $\mathcal{F}$ (D) \\\hline\hline,late after line=\\\hline,davisnames}

\begin{table}[h]\centering
\resizebox{\columnwidth}{!}{
\begin{filecontents*}{csv/w9atwssz_cs120_cst120_exceptbiggest_davis16_val.csv}
Sequence,J(M),J(O),J(D),F(M),F(O),F(D)
blackswan,0.468,0.333,-0.058,0.548,0.938,-0.008
bmx-trees,0.596,0.744,0.199,0.787,0.923,0.087
breakdance,0.758,1.000,0.019,0.763,1.000,0.010
camel,0.812,1.000,0.138,0.771,1.000,0.136
car-roundabout,0.926,1.000,0.006,0.816,1.000,-0.033
car-shadow,0.897,1.000,0.014,0.848,1.000,-0.020
cows,0.812,0.990,-0.012,0.716,0.961,-0.012
dance-twirl,0.748,1.000,0.021,0.758,1.000,0.104
dog,0.680,0.845,-0.287,0.575,0.655,-0.207
drift-chicane,0.608,0.680,-0.178,0.699,0.760,-0.055
drift-straight,0.883,1.000,0.065,0.825,0.979,0.290
goat,0.137,0.000,-0.157,0.242,0.000,-0.137
horsejump-high,0.841,1.000,0.060,0.892,1.000,-0.001
kite-surf,0.484,0.458,0.192,0.465,0.250,-0.003
libby,0.745,0.957,0.191,0.849,1.000,0.109
motocross-jump,0.682,0.763,0.142,0.527,0.579,0.176
paragliding-launch,0.615,0.667,0.283,0.275,0.051,0.274
parkour,0.703,0.816,-0.028,0.790,0.980,0.096
scooter-black,0.849,1.000,-0.044,0.716,1.000,0.062
soapbox,0.885,1.000,0.032,0.847,1.000,0.043
Average,0.707,0.813,0.030,0.686,0.804,0.046


\end{filecontents*}
\csvreader[Rta]{csv/w9atwssz_cs120_cst120_exceptbiggest_davis16_val.csv}{}{\Sequence & \JM & \JO & \JD & \FM & \FO & \FD }}
\vspace{0.10cm}
\caption{Results given for every video of DAVIS2016 dataset. Reported scores per video are the average Jaccard score over frames in the video. The very last row is the average over videos scores.
$\mathcal{J}$ is the Jaccard index and $\mathcal{F}$ is the Countour Accuracy. The Mean ($M$) is the average of the scores, the Recall ($O$) is the fraction of frames per video with a score higher than $0.5$, and the Decay ($D$) is the degradation of the score over time in the video.
}
\end{table}

\subsection{Detailed results for DAVIS2017-motion}

\csvnames{davis17names}{1=\Sequence,2=\JMean,3=\FMean}
\csvstyle{Rte}{tabular=|l|c|c|,table head=\hline Sequence & J-Mean & F-Mean \\\hline\hline,late after line=\\\hline,davis17names}
\begin{table}\centering
\begin{filecontents*}{csv/2r9flv5x-D17-Sequence-paper.csv}
Sequence,JMean,FMean
bike-packing\_1,0.078,0.327
bike-packing\_2,0.254,0.267
blackswan\_1,0.477,0.575
bmx-trees\_1,0.608,0.804
breakdance\_1,0.450,0.526
camel\_1,0.772,0.726
car-roundabout\_1,0.913,0.810
car-shadow\_1,0.896,0.845
cows\_1,0.821,0.728
dance-twirl\_1,0.444,0.576
dog\_1,0.654,0.571
dogs-jump\_1,0.019,0.147
dogs-jump\_2,0.254,0.323
dogs-jump\_3,0.303,0.351
drift-chicane\_1,0.557,0.627
drift-straight\_1,0.886,0.829
goat\_1,0.220,0.300
gold-fish\_1,0.018,0.240
gold-fish\_2,0.282,0.336
gold-fish\_3,0.357,0.356
gold-fish\_4,0.000,0.000
gold-fish\_5,0.000,0.000
horsejump-high\_1,0.721,0.794
india\_1,0.066,0.085
india\_2,0.163,0.172
india\_3,0.072,0.108
judo\_1,0.225,0.397
judo\_2,0.286,0.414
kite-surf\_1,0.449,0.464
lab-coat\_1,0.332,0.350
libby\_1,0.746,0.853
loading\_1,0.057,0.246
loading\_2,0.128,0.259
loading\_3,0.427,0.514
mbike-trick\_1,0.444,0.436
motocross-jump\_1,0.446,0.435
paragliding-launch\_1,0.577,0.286
parkour\_1,0.531,0.646
pigs\_1,0.105,0.433
pigs\_2,0.437,0.431
pigs\_3,0.059,0.217
scooter-black\_1,0.843,0.724
shooting\_1,0.434,0.568
soapbox\_1,0.487,0.696
\end{filecontents*}
\csvreader[Rte]{csv/2r9flv5x-D17-Sequence-paper.csv}{}{\Sequence & \JMean & \FMean}
\end{table}

\begin{table}[!ht]
    \centering
    \resizebox{\columnwidth}{!}{
    \begin{tabular}{|l|l|l|l|l|l|l|}
    \hline
        J\&FMean & JMean & JRecall & JDecay & FMean & FRecall & FDecay \\ \hline
        0.422 & 0.393 & 0.387 & 0.004 & 0.450 & 0.437 & 0.024 \\ \hline
    \end{tabular}}
    \vspace{0.10cm}
\caption{Results given for every video of DAVIS2017-motion dataset. The very last row is the average score over all the videos for the different criteria.}
\end{table}


\small
\vspace{-0.15cm}
\begin{thebibliography}{28}
\providecommand{\natexlab}[1]{#1}
\providecommand{\url}[1]{\texttt{#1}}
\expandafter\ifx\csname urlstyle\endcsname\relax
  \providecommand{\doi}[1]{doi: #1}\else
  \providecommand{\doi}{doi: \begingroup \urlstyle{rm}\Url}\fi




\bibitem{batchelor1967}
G.K. Batchelor.
\newblock \emph{An introduction to fluid dynamics}.
\newblock Cambridge University Press, 1967.

\bibitem{bideau_moa-net_2018}
P. Bideau, R. Menon, and E. Learned-Miller.
\newblock MoA-Net: Self-supervised motion segmentation.
\newblock In \emph{European Conference on Computer Vision Workshops (ECCVW)}, 2018.


\bibitem{butler2012}
D.J. Butler, J. Wulff, G.B. Stanley, and M.J. Black. 
\newblock A naturalistic open source movie for optical flow evaluation,
\newblock In \emph{European Conference on Computer Vision (ECCV)}, 2012.

\bibitem{caron2021}
M. Caron, H. Touvron, I. Misra, H. Jégou, J. Mairal, P. Bojanowski, and A. Joulin.
\newblock Emerging properties in self-supervised vision transformers.
\newblock In \emph{Int. Conf. on Computer Vision
(ICCV)}, Oct. 2021.

\bibitem{cheng2021}
B. Cheng, A. G. Schwing, and A. Kirillov.
\newblock Per-pixel classification is not all you need for semantic segmentation.
\newblock In \emph{Conference on Neural Information Processing Systems (NeurIPS)}, 2021.

\bibitem{cheng2017}
J. Cheng, Y.-H. Tsai, S. Wang, and M.-H. Yang.
\newblock Segflow: Joint learning for video object segmentation and optical flow.
\newblock In \emph{International Conference on Computer Vision (ICCV)}, Venice, 2017.

\bibitem{choudhury2022}
S. Choudhury, L. Karazija, I. Laina, A. Vedaldi, and C. Rupprecht.
\newblock Guess what moves: Unsupervised video and image segmentation by anticipating motion.
\newblock \emph{British Machine Vision Conf. (BMVC)}, 2022.

\bibitem{cremers2005}
D. Cremers and S. Soatto.
\newblock Motion competition: A variational approach to piecewise parametric motion segmentation.
\newblock \emph{International Journal of Computer Vision}, 62(3):249–265, 2005.


\bibitem{dave2019}
A. Dave, P. Tokmakov, and D. Ramanan.
\newblock Towards segmenting anything that moves. 
\newblock In \emph{Int. Conference on Computer Vision Workhops (ICCVW)}, Seoul, 2019.

\bibitem{ding2022}
S. Ding, W. Xie, Y. Chen, R. Qian, X. Zhang, H. Xiong, and Q. Tian.
\newblock Motion-inductive self-supervised object discovery in videos.
\newblock \emph{arxiv.org/abs/2210.00221}, October 2022.

\bibitem{duke2021}
B. Duke, A. Ahmed, C. Wolf, P. Aarabi, and G. W. Taylor.
\newblock {SSTVOS}: Sparse spatiotemporal transformers for video object segmentation.
\newblock In \emph{Conf. on Computer Vision and Pattern Recognition (CVPR)}, 2021.




\bibitem{griffin_tukey-inspired_2019}
B. Griffin and J. Corso.
\newblock Tukey-inspired video object segmentation.
\newblock In \emph{IEEE Winter Conference on Applications of Computer Vision (WACV)}, Waikoloa Village, January 2019.

\bibitem{haller2021}
E. Haller, A. M. Florea, and M. Leordeanu. \newblock Iterative knowledge exchange between deep learning and space-time spectral clustering for unsupervised segmentation in videos.
\newblock \emph{IEEE Transactions on Pattern Analysis and Machine Intelligence}, 44(11):7638–7656, 2021.

\bibitem{han2023}
K. Han, Y. Wang, H. Chen, X. Chen, J. Guo, Z. Liu, Y. Tang,
A. Xiao, C. Xu, Y. Xu, Z. Yang, Y. Zhang, and D. Tao
\newblock A survey on vision transformer.
\newblock \emph{IEEE Transactions on Pattern Analysis and Machine Intelligence}, 45(1):87-110, January 2023.

\bibitem{irani1998}
M. Irani and P. Anandan.
\newblock A unified approach to moving object detection in 2D
and 3D scenes.
\newblock \emph{IEEE Transactions on Pattern Analysis and Machine Intelligence}, 20(6):577–589, June 1998.

\bibitem{jain_fusionseg_2017}
S.D. Jain, B. Xiong, and K. Grauman.
\newblock FusionSeg: Learning to combine motion and appearance for fully automatic segmentation of generic objects in videos.
\newblock In \emph{Conference on Computer Vision and Pattern Recognition (CVPR)}, Honolulu, 2017.


\bibitem{kingma2019}
D.P. Kingma and M. Welling.
\newblock An introduction to variational autoencoders.
\newblock \emph{Foundations and Trends in Machine Learning}, 12(4):307-392, 2019.


\bibitem{lamdouar_camouflage_2020}
H. Lamdouar, C. Yang, W. Xie, and A. Zisserman.
\newblock Betrayed by motion: Camouflaged object discovery via motion segmentation.
\newblock In \emph{Asian Conf. on Computer Vision (ACCV)}, Kyoto, 2020.


\bibitem{lao2023}
D. Lao, Z. Hu, F. Locatello, Y. Yang, and S. Soatto.
\newblock Divided attention: Unsupervised multi-object discovery with contextually separated slots.
\newblock \emph{arXiv:2304.01430}, April 2023.

\bibitem{li2013}
F. Li, T. Kim, A. Humayun, D. Tsai, and J. M. Rehg.
\newblock Video segmentation by tracking many figure-ground segments.
\newblock In \emph{International Conference on Computer Vision (ICCV)}, Sydney, 2013.


\bibitem{lian2023}
L. Lian, Z. Wu, S. X. Yu.
\newblock Bootstrapping objectness from videos by relaxed common fate and visual grouping
\newblock \emph{Conference on Computer Vision and Pattern Recognition}, Vancouver, June 2023.

\bibitem{lbgfs89}
D. Liu and J. Nocedal.
\newblock On the limited memory BFGS method for large scale optimization.
\newblock In \emph{Mathematical Programming}, 45(1-3):503-528, 1989.

\bibitem{liu2021}
R. Liu, Z. Wu, S. X. Yu, and S. Lin.
\newblock The Emergence of objectness: Learning zero-shot segmentation from videos.
\newblock In \emph{Conference on Neural Information Processing Systems (NeurIPS)}, 2021.

\bibitem{locatello2020}
F. Locatello, D. Weissenborn, T. Unterthiner, A. Mahendran,
G. Heigold, J. Uszkoreit, A. Dosovitskiy, and T. Kipf.
\newblock Object-centric learning with slot attention.
\newblock In \emph{Conference on Neural Information Processing Systems (NeuRIPS)}, 2020.



\bibitem{mahendran2018}
A. Mahendran, J. Thewlis, and A. Vedaldi.
\newblock Self-supervised segmentation by grouping optical flow.
\newblock In \emph{European Conf. on Computer Vision Workshops (ECCVW)}, Munich, 2018.

\bibitem{mattheus2020}
J. Mattheus, H. Grobler, and A. M. Abu-Mahfouzl.
\newblock A review of motion segmentation: Approaches and major challenges.
\newblock \emph{International Multidisciplinary Information Technology and Engineering Conference (IMITEC)}, November 2020.

\bibitem{ft3d2016}
N. Mayer, E. Ilg, P. H{\"a}usser, P. Fischer, D. Cremers, A. Dosovitskiy and T. Brox.
\newblock A large dataset to train convolutional networks for disparity, optical flow, and scene flow estimation.
\newblock In \emph{Conf. on Computer Vision and Pattern Recognition (CVPR)}, June 2016.

\bibitem{meunier2021}
E. Meunier and P. Bouthemy.
\newblock Unsupervised computation of salient motion maps
from the interpretation of a frame-based classification network, \newblock In \emph{British Machine Vision Conference (BMVC)}, 2021.

\bibitem{meunier2022}
E. Meunier, A. Badoual and P. Bouthemy.
\newblock EM-driven unsupervised learning for efficient motion segmentation.
\newblock In \emph{IEEE Trans. on Pattern Anal. and Machine Intelligence}, 45(4):4462-4473, April 2023.

\bibitem{meunier2023-cvpr}
E. Meunier and P. Bouthemy.
\newblock Unsupervised space-time network for temporally-consistent segmentation of multiple motions. 
\newblock In \emph{Conf. on Computer Vision and Pattern Recognition (CVPR)}, 2023.

\bibitem{mitiche1996}
A. Mitiche and P. Bouthemy.
\newblock Computation and analysis of image motion: A synopsis of current
problems and methods.
\newblock \emph{International Journal of Computer Vision}, 19(1):29-55, 1996.


\bibitem{ochs2014}
P. Ochs, J. Malik, and T. Brox.
\newblock Segmentation of moving objects by long term video analysis.
\newblock \emph{IEEE Transactions on Pattern Analysis and Machine Intelligence}, 36(6):1187-1200, June~2014.

\bibitem{odobez1995}
J.-M. Odobez and P.~Bouthemy.
\newblock {MRF}-based motion segmentation exploiting a 2{D} motion model robust estimation.
\newblock In \emph{International Conference on Image Processing (ICIP)}, Washington, October 1995.




\bibitem{papazoglou_fast_2013}
A. Papazoglou and V. Ferrari.
\newblock Fast object segmentation in unconstrained video.
\newblock In \emph{IEEE International Conference on Computer Vision (ICCV)}, Sydney, December 2013.

\bibitem{perazzi_benchmark_2016}
F. Perazzi, J. Pont-Tuset, B. Mc Williams, L. Van~Gool, M. Gross, and A. Sorkine-Hornung.
\newblock A benchmark dataset and evaluation methodology for video object segmentation.
\newblock In \emph{Conf. on Computer Vision and Pattern Recognition (CVPR)}, Las Vegas, June 2016.

\bibitem{ponimatkin2023}
G. Ponimatkin, N. Samet, Y. Xiao, Y. Du, R. Marlet, and V. Lepetit.
\newblock A simple and powerful global optimization for unsupervised video object segmentation.
\newblock In \emph{Winter Conference on Applications of Computer Vision (WACV)}, Waikoloa, Jan. 2023.

\bibitem{davis2017}
J. Pont-Tuset, F. Perazzi, S. Caelles, P. Arbeláez, A. Sorkine-Hornung, and L. Van Gool.
\newblock The 2017 Davis challenge on video object segmentation.
\newblock \emph{arXiv:1704.00675}, 2017.

\bibitem{ranjan_competitive_2019}
A. Ranjan, V. Jampani, L. Balles, K. Kim, D. Sun, J. Wulff, and M.J. Black.
\newblock Competitive collaboration: Joint unsupervised learning of depth, camera motion, optical flow and motion segmentation.
\newblock In \emph{Conf. on Computer Vision and Pattern Recognition (CVPR)}, 2019.

\bibitem{ronneberger_Unet_2015}
O. Ronneberger, P. Fischer, and T. Brox.
\newblock U-net: Convolutional networks for biomedical image segmentation.
\newblock In \emph{Int. Conf. on Medical Image Computing and Computer Assisted Intervention (MICCAI)}, Munich, October 2015.


\bibitem{sun2012}
D. Sun, E. B. Sudderth, and M. J. Black.
\newblock Layered segmentation and optical flow estimation over time.
\newblock In \emph{Conference on Computer Vision and Pattern Recognition (CVPR)}, Providence, June 2012. 

\bibitem{teed_raft_2020}
Z. Teed and J. Deng.
\newblock RAFT: Recurrent all-pairs field transforms for optical flow.
\newblock In \emph{European Conf. on Computer Vision (ECCV)}, 2020.

\bibitem{tokmakov_learning_2017}
P. Tokmakov, K. Alahari, and C. Schmid.
\newblock Learning motion patterns in videos.
\newblock In \emph{Conf. on Comp. Vision and Pattern Recog. (CVPR)}, 2017.


\bibitem{tu2019}
Z. Tu, W. Xie, D. Zhang, R. Poppe, R. C. Veltkamp, B. Li, and J. Yuan.
\newblock A survey of variational and CNN-based optical flow techniques.
\newblock \emph{Signal Proc.: Image Communication}, 72:9-24, March 2019.

\bibitem{unser1999}
M. Unser.
\newblock Splines: A perfect fit for signal and image processing.
\newblock \emph{Signal Processing Magazine}, 16(6):22-38, November 1999.



\bibitem{wang1994}
J.Y.A. Wang and E.H. Adelson.
\newblock Representing moving images with layers.
\newblock \emph{IEEE Trans. on Image Processing}, 3(5):625-638, Sept.1994.


\bibitem{xiao-shah2005}
J. Xiao and M. Shah.
\newblock Motion layer extraction in the presence of occlusion using graph cuts.
\newblock \emph{IEEE Transactions on Pattern Analysis and Machine Intelligence}, 27(10):1644-1659, October 2005.

\bibitem{xie2022}
J. Xie, W. Xie, and A. Zisserman.
\newblock Segmenting moving objects via an object-centric layered representation.
\newblock In \emph{Conference on Neural Information Processing Systems (NeurIPS)}, 2022.

\bibitem{xu2021}
X. Xu, L. Zhang, L.-F. Cheong, Z. Li, and C. Zhu.
\newblock Learning clustering for motion segmentation.
\newblock \emph{IEEE Transactions on Circuits and Systems for Video Technology}, 32(3):908-919, March 2022.

\bibitem{yang_motion-grouping_2021}
C. Yang, H. Lamdouar, E. Lu, A. Zisserman, and W. Xie.
\newblock Self-supervised video object segmentation by motion grouping.
\newblock In \emph{International Conference on Computer Vision (ICCV)}, Oct. 2021.

\bibitem{yang_unsupervised_2019}
Y. Yang, A. Loquercio, D. Scaramuzza, and S. Soatto.
\newblock Unsupervised moving object detection via contextual information separation.
\newblock In \emph{Conf. on Computer Vision and Pattern Recog. (CVPR)}, 2019.
  
\bibitem{dystab2021}
Y. Yang, B. Lai, and S. Soatto, 
\newblock DyStaB: Unsupervised object segmentation via dynamic-static bootstrapping.
\newblock In \emph{Conference on Computer Vision and Pattern Recognition (CVPR)}, 2021.

\bibitem{ye2022}
V. Ye, Z. Li, R. Tucker, A. Kanazawa, and N. Snavely.
\newblock Deformable sprites for unsupervised video decomposition.
\newblock In \emph{Conf. on Computer Vision and Pattern Recognition (CVPR)}, New Orleans, 2022.


\bibitem{zappella2008}
L. Zappella, X. Llado, and J. Salvi
\newblock Motion segmentation: A review.
\newblock \emph{Frontiers in Artificial Intelligence and Applic.}, 184:398-407, Jan. 2008.


\bibitem{wang_survey-VOS}
T. Zhou, F. Porikli, D.J. Crandall, L. Van Gool, and W. Wang.
\newblock A survey on deep learning technique for video segmentation.
\newblock \emph{IEEE Trans. on Pattern Anal. and Mach. Intelligence}, 45(6):7099-7122, 2023.

\end{thebibliography}

\begin{thebibliography}{28}
\providecommand{\natexlab}[1]{#1}
\providecommand{\url}[1]{\texttt{#1}}
\expandafter\ifx\csname urlstyle\endcsname\relax
  \providecommand{\doi}[1]{doi: #1}\else
  \providecommand{\doi}{doi: \begingroup \urlstyle{rm}\Url}\fi



\bibitem{lao2023}
D. Lao, Z. Hu, F. Locatello, Y. Yang, and S. Soatto.
\newblock Divided attention: Unsupervised multi-object discovery with contextually separated slots.
\newblock \emph{arXiv:2304.01430}, April 2023.


\bibitem{meunier2023-cvpr}
E. Meunier and P. Bouthemy.
\newblock Unsupervised space-time network for temporally-consistent segmentation of multiple motions. 
\newblock In \emph{Conference on Computer Vision and Pattern Recognition (CVPR)}, Vancouver, June 2023.


\end{thebibliography}
\end{document}